\documentclass[10pt,journal,compsoc]{IEEEtran}
\ifCLASSOPTIONcompsoc
  \usepackage[nocompress]{cite}
\else
  \usepackage{cite}
\fi

\ifCLASSINFOpdf
   \usepackage[pdftex]{graphicx}
   \usepackage{epstopdf}
   \graphicspath{{../fig/}{../jpeg/}}
   \DeclareGraphicsExtensions{.pdf,.jpeg,.png,.eps}
\else
   \usepackage[dvips]{graphicx}
   \graphicspath{{../fig/}}
   \DeclareGraphicsExtensions{.eps, .png, .pdf}
\fi

\usepackage{amsmath}
\usepackage{amsfonts}

\usepackage{array}
\usepackage{tabularx}
\usepackage{multirow}
\usepackage{ctable} % for \specialrule command

\usepackage{float}
\ifCLASSOPTIONcompsoc
  \usepackage[caption=false,font=footnotesize,labelfont=sf,textfont=sf]{subfig}
\else
  \usepackage[caption=false,font=footnotesize]{subfig}
\fi

\def\etal{\emph{et al.}}
\def\eg{\emph{e.g.}}
\def\C{\mathbb{C}}
\newcolumntype{L}[1]{>{\raggedright\arraybackslash}p{#1}}
\newcolumntype{C}[1]{>{\centering\arraybackslash}p{#1}}
\newcolumntype{R}[1]{>{\raggedleft\arraybackslash}p{#1}}

\begin{document}

\title{Confidence Propagation through CNNs for Guided Sparse Depth Regression}

\author{Abdelrahman~Eldesokey,~\IEEEmembership{Student~Member,~IEEE,}
        Michael~Felsberg,~\IEEEmembership{Senior~Member,~IEEE,}
        and~Fahad~Shahbaz~Khan,~\IEEEmembership{Member,~IEEE}% <-this % stops a space
\IEEEcompsocitemizethanks{\IEEEcompsocthanksitem All authors are with the Department of Electrical Engineering, Link\"oping University, SE-581 83 Link\"oping, Sweden
.\protect\\
E-mail: abdelrahman.eldesokey@liu.se
\IEEEcompsocthanksitem Fahad Khan is also with the Inception Institute of Artificial
Intelligence Abu Dhabi, UAE}% <-this % stops an unwanted space
}

\markboth{Journal of \LaTeX\ Class Files,~Vol.~14, No.~8, August~2015}%
{Shell \MakeLowercase{\textit{et al.}}: Bare Demo of IEEEtran.cls for Computer Society Journals}

\IEEEtitleabstractindextext{%
\begin{abstract}
Generally, convolutional neural networks (CNNs) process data on a regular grid, \textit{e.g.} data generated by ordinary cameras. Designing CNNs for sparse and irregularly spaced input data is still an open research problem with numerous applications in autonomous driving, robotics, and surveillance. In this paper, we propose an algebraically-constrained normalized convolution layer for CNNs with highly sparse input that has a smaller number of network parameters compared to related work. We propose novel strategies for determining the confidence from the convolution operation and propagating it to consecutive layers. We also propose an objective function that simultaneously minimizes the data error while maximizing the output confidence. To integrate structural information, we also investigate fusion strategies to combine depth and RGB information in our normalized convolution network framework. In addition, we introduce the use of output confidence as an auxiliary information to improve the results. The capabilities of our normalized convolution network framework are demonstrated for the problem of scene depth completion. Comprehensive experiments are performed on the KITTI-Depth %benchmark and the results clearly demonstrate that the proposed approach achieves superior performance while requiring only about 5\% of the number of parameters compared to the state-of-the-art methods.
and the NYU-Depth-v2 datasets. The results clearly demonstrate that the proposed approach achieves superior performance while requiring only about 1-5\% of the number of parameters compared to the state-of-the-art methods.

\end{abstract}

}

\maketitle

\IEEEdisplaynontitleabstractindextext

% \fi
\IEEEpeerreviewmaketitle

\IEEEraisesectionheading{\section{Introduction}\label{sec:introduction}}
\IEEEPARstart{S}{ensors} with dense output such as monochrome, color, and thermal cameras have been extensively exploited by machine learning methods in many computer vision applications. Images generated by these sensors are typically fully dense due to their passive nature and different image regions are initially equally relevant to the machine learning algorithms. However, other, mostly active, sensors such as ToF cameras, LiDAR, RGB-D, and event cameras produce sparse output. This sparsity is usually caused by their active sensing, which leaves many data regions empty. The sparse output from these sensors imposes fundamental challenges on the machine learning methods as data relevance is not uniform and further processing is required to either reconstruct or ignore these missing regions.

The degree of sparsity and data pattern differ from one sensor to another and machine learning methods should be able to handle different scenarios. Handling sparsity would open up for numerous applications in robotics, autonomous driving, and surveillance due to the depth information made available by active sensors. Therefore, a major task is \emph{scene depth completion}, which aims to reconstruct a dense depth map from the sparse output produced by active depth sensors. Scene depth completion is crucial for tasks that require situation awareness for decision support. Besides, the availability of a reliability measure, \emph{i.e.} confidence, is also desirable since it gives an indication about the trustworthiness of the output. A confidence measure would be highly beneficial for safety and decision making applications such as obstacle detection and avoidance in autonomous driving.

\setlength{\tabcolsep}{1pt}
\begin{figure*}[t]
\includegraphics[width=\textwidth]{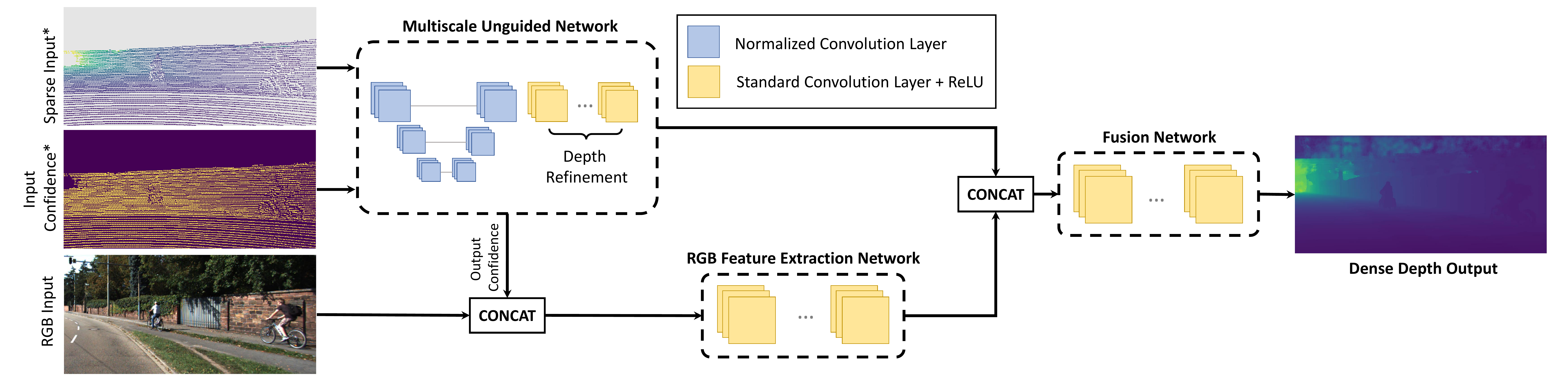}
\caption{Our scene depth completion pipeline on an example image from the KITTI-Depth dataset \cite{Uhrig2017}. The input to the pipeline is a very sparse projected LiDAR point cloud, an input confidence map which has zeros at missing pixels and ones otherwise, and an RGB image. The sparse point cloud input and the input confidence are fed to a multi-scale unguided network that acts as a generic estimator for the data. Afterwards, the continuous output confidence map is concatenated with the RGB image and fed to a feature extraction network. The output from the unguided network and the RGB feature extraction networks are concatenated and fed to a fusion network which produces the final dense depth map. [*Images were dilated for visual clarity ]}
\label{fig:intro}
\end{figure*}

A key problem in scene depth completion is the identification of missing values and distinguishing them from regions with zero values. One way to identify missing data is using binary validity masks with zeros at regions with missing values and ones otherwise. Validity masks have been extensively used in the literature \cite{ren2015shepard, Uhrig2017, Chodosh2018, Liu2018, ncnn, hms} to inform the learning method about the missing regions in the data. However, validity masks suffer from saturation in multi-stage learning such as Convolutional Neural Networks (CNNs) as shown by \cite{valeo}. Instead, the saturation problem can be avoided by treating the binary validity masks as continuous confidence fields describing the reliability of the data. Additionally, this enables confidence propagation, which helps to keep track of the reliability of the data throughout the processing pipeline.

Most recent works for solving the scene depth completion task are based on CNNs, and show a great success \cite{mit,valeo,hms,Uhrig2017,ncnn,Chodosh2018}. Typically, a deep CNN is trained to construct a dense depth map given either a sparse depth input only or sparse depth aside with an RGB image. The former case is denoted as \emph{unguided} depth completion, while the latter is called \emph{guided} depth completion. The role of the network in both cases is to learn the manifold where the data live in. Since the publicly available datasets for scene depth completion such as the KITTI-Depth dataset \cite{Uhrig2017} have a very high spatial resolution, the state-of-the-art methods \cite{mit, valeo, hms} demand huge CNNs with millions of parameters to solve the problem. Unfortunately, this hinders the deployment of such methods in autonomous driving and robotic systems with limited computational and memory resources. It is desirable to design compact CNN architectures, while propagating confidences, for real-world applications with limited resources.

In this paper, we introduce the normalized convolution layer, which allows performing unguided scene depth completion on highly sparse data with a smaller number of parameters than related methods. Our proposed method treats the validity masks as a continuous confidence field and we propose a new criteria to propagate confidences between CNN layers, thereby allowing us to produce a point-wise continuous confidence map for the output from the network. Furthermore, we algebraically constrain the network learned filters to be non-negative, acting as a weighting function for the neighborhood. This allows the network to converge faster and achieves remarkably better results. We also propose a loss function that aims to simultaneously minimize the data error and maximize the output confidence.

%% modified by fahad
The proposed normalized convolution network generally performs well on smooth surfaces when using only the depth information. However, it performs less well across edges due to lack of structural information. This can be mitigated by using guidance from RGB images in order to integrate useful structural information into our network. Both RGB and depth information can be fused in our proposed framework in multiple ways. In this work, we investigate both early and late fusion schemes in two state-of-the-art architectures. The first is a multi-stream architecture inspired by \cite{ncnn} which is highly relevant to our proposed method and the second is an encoder-decoder architecture with skip-connections inspired by\cite{Liu2018,Ronneberger2015}. In addition, we introduce the use of output confidences as guidance information aside with the RGB images. On the KITTI-Depth \cite{Uhrig2017} \textcolor{black}{and the NYU-Depth-v2 \cite{nyu} datasets}, our proposed method achieves state-of-the-art results while requiring a significantly lower number of parameters \textcolor{black}{($\sim356$k and $\sim484$k parameters) respectively,} compared to all the existing state-of-the-art methods (with millions of parameters). This proves the efficiency of our proposed method.  An illustration of the proposed pipeline is shown in Figure \ref{fig:intro}.

%The proposed normalized convolution network generally performs well on smooth surfaces when using only the depth information. However, it performs less well across edges due to lack of structural information. This can be mitigated by using guidance from RGB images in order to integrate useful structural information into our network. Both RGB and depth information can be fused in our proposed framework in multiple ways. In this work, we investigate both early and late fusion schemes in two state-of-the-art architectures. The first is a multi-stream architecture inspired by \cite{ncnn} which is highly relevant to our proposed method and the second is an encoder-decoder architecture with skip-connections inspired by\cite{Liu2018,Ronneberger2015}. In addition, we introduce the use of output confidences as guidance information aside with the RGB images. On the KITTI-Depth benchmark, our proposed multi-stream architecture with late fusion achieves remarkable results compared to published state-of-the-art methods while requiring significantly lower number of parameters than comparable methods. This demonstrates the efficiency of our proposed method, which eliminated the need for a huge number of parameters to achieve state-of-the-art results. Furthermore, we demonstrate that incorporating output confidence as guidance information led to an improvement in the results.  An illustration of the proposed pipeline is shown in Figure \ref{fig:intro}.

The rest of the paper is organized as follows. Section \ref{sec:related} gives an overview of the relevant work in the literature. In section \ref{sec:nconv}, we describe the normalized convolution framework in details. Section \ref{sec:ncnn} describes our early work on unguided depth completion in \cite{bmvc}. Section \ref{sec:gncnn} introduces our proposed approach to fuse sparse depth, RGB images, and the output confidences. \textcolor{black}{Extensive experiments on both our prior work \cite{bmvc} and the proposed fusion schemes are presented in section \ref{sec:exp}. Finally, we provide a thorough analysis for our proposed method in section \ref{sec:analys}}. The conclusion is given in section \ref{sec:concl}.

\section{Related Work} \label{sec:related}

Scene depth completion has become a fundamental task in computer vision ever since the emergence of active sensors with depth capabilities. Generally, it was treated as a hole-filling or inpainting problem using classical image processing methods \cite{6335696,7003768,6865733}. Recently, with the advent of deep learning, specifically Convolutional Neural Networks (CNNs), scene depth completion has matured into a separate task than inpainting. Typically, scene depth completion is performed on depth maps with optional  guidance from RGB images. Differently, inpainting is mostly performed on RGB or grayscale images. In addition, the objective of scene depth completion is to minimize some error measure such as the \textit{L1} or the \textit{L2} norm, while inpainting aims also to provide realistic output \cite{Liu2018}.

For the task of unguided scene depth completion, where the input is only the depth map, Chodosh \etal \cite{Chodosh2018} utilized compressed sensing to handle the sparsity, while using a binary mask to filter out the missing values. Ma \etal \cite{mit} utilized an encoder-decoder architecture with self-supervised framework to predict the dense output. Uhrig \etal \cite{Uhrig2017} proposed a sparsity-invariant convolution layer that utilizes binary validity masks to normalize the sparse input. The proposed layer was used to train a network with a sparse depth map aside with a binary validity mask as input and a dense depth map as output. Similarly, \cite{hms} utilized the sparsity-invariant layer in more complex CNN architectures. Hua and Gong \cite{ncnn} proposed a similar layer, which uses the trained convolution filter to normalize the sparse input. Contrarily, Jaritz \etal \cite{valeo} compared different architectures and argued that the use of validity masks degrades the performance due to the saturation of the masks at early layers within the CNN. This effect is avoided by the use of continuous confidences as proposed in our prior work on unguided depth completion \cite{bmvc}.

Due to the sub-optimal performance of unguided methods across edges, several recent approaches urged to use guidance from auxiliary data such as RGB images or surface normals. Ma \etal \cite{mit} used an early fusion scheme to combine sparse depth input with the corresponding RGB image, which was demonstrated to perform very well. On the other hand, Jaritz \etal \cite{valeo} argued that late-fusion performs better with their proposed architecture, which was also demonstrated in \cite{ncnn}. Wirges \etal \cite{7995866} used a combination of RGB images and surface normals to guide the process of depth upsampling. Konno \etal \cite{konno2015intensity} utilized a residual interpolation method to combine a low-resolution depth map with a high resolution RGB image to produce a high resolution depth map.

%The proposed normalized convolution network generally performs well on smooth surfaces when using only the depth information. However, it performs less well across edges due to lack of structural information. This can be mitigated by using guidance from RGB images in order to integrate useful structural information into our network. Both RGB and depth information can be fused in our proposed framework in multiple ways.

%% Modified by fahad
Different to the aforementioned approaches, we proposed a normalized convolution layer in \cite{bmvc}, which takes in a sparse input aside with a continuous confidence map to perform unguided scene depth completion. Different to \cite{ncnn}, we impose algebraic constraints on the trained filters to be non-negative, which allow the network to converge faster while requiring significantly lower number of parameters. Further, we derive a confidence propagation criteria between layers, which enables producing a point-wise continuous confidence map aside with the dense output from the CNN. As an extention to our priod work \cite{bmvc}, we use guidance from RGB images, and Different to \cite{mit,valeo,hms}, we also use guidance from the output confidence produced by the unguided network. Our results clearly show that using guidance from the output confidence leads to a significant improvement in performance. Our proposed multi-stream architecture with late fusion achieves remarkable results compared to published state-of-the-art methods while requiring significantly lower number of parameters than comparable methods. This demonstrates the efficiency of our proposed method, which eliminated the need for a huge number of parameters to achieve state-of-the-art results.

\section{Normalized Convolution} \label{sec:nconv}
The concept of Normalized Convolution was first introduced by Knutsson and Westin \cite{Knutsson} based on the theory of confidence-equipped signals. Assume a discrete finite signal $\mathit{f}$ that is periodically sampled. \textcolor{black}{This signal $\mathit{f}$ could, \eg, depict a sparse depth field.} At each sample point, the neighborhood is finite and can be represented as a vector $\mathbf{f} \in \C^n$. This signal is accompanied by a confidence field $\mathit{c}$, which describes the reliability of each sample value. Confidences are typically non-negative, \textcolor{black}{and in the case of the sparse depth field, zero confidence indicates the absence of the corresponding depth sample value}. The confidence field $\mathit{c}$ is sampled in the same manner as $\mathit{f}$ and the confidence of each neighborhood is represented as a finite vector $\mathbf{c} \in \C^n$.

In \cite{Knutsson}, the signal is modeled locally by projecting each sample $\mathbf{f}$ onto a subspace spanned by some basis functions, \textit{e.g.} polynomials, complex exponentials, or the na\"ive basis. The set of basis functions $ \{ \mathbf{b}_i \in \C^n \}_1^m $ have the same dimensionality as the sample $\mathbf{f}$ and its corresponding confidence $\mathbf{c}$. The sample $\mathbf{f}$ is expressed with respect to the basis functions  
%\begin{equation}\label{eq:1}
%\mathbf{f} = r_1 \mathbf{b}_1 + r_2 \mathbf{b}_2 + \dots + r_m \mathbf{b}_m \enspace,
%\end{equation}
%where $r_1 \dots r_m$ are the coordinates of the sample $\mathbf{f}$ in the subspace spanned by the basis functions. The basis functions can be
arranged into the columns of a $\mathit{n} \times \mathit{m}$ matrix $\mathbf{B}$ as:
\begin{equation}\label{eq:1}
\mathbf{f} = \mathbf{B} \mathbf{r}
\enspace,
\end{equation}
where $\mathbf{r}$ holds the coordinates of the sample $\mathbf{f}$ with respect to the basis functions.

Finding these coordinates is usually formulated as a weighted least-squares problem:
\begin{equation} \label{eq:2}
\mathrm{arg} \min_{\mathbf{r} \in \C^n} || \mathbf{B} \mathbf{r} - \mathbf{f} ||_\mathbf{W}	\enspace,
\end{equation}
\textcolor{black}{where $\mathbf{W}$ is the weight matrix for the least-squares problem. In our case, the confidence $\mathbf{c}$ is used to weight the sample $\mathbf{f}$ and an applicability function $\mathbf{a} \in \C^n$, acts as a weight for the basis.} The solution $\mathbf{\hat{r}}$ to this weighted least-squares problem reads:
\begin{equation} \label{eq:3}
\begin{split}
\mathbf{\hat{r}} &=  (\mathbf{B}^* \mathbf{W} \mathbf{B})^{-1} \mathbf{B}^* \mathbf{W}\mathbf{f} \enspace, \\ 
&= (\mathbf{B}^* \mathbf{D}_\mathbf{a} \mathbf{D}_\mathbf{c} \mathbf{B})^{-1} \mathbf{B}^* \mathbf{D}_\mathbf{a} \mathbf{D}_\mathbf{c} \mathbf{f} \enspace,
\end{split}
\end{equation}
where $\mathbf{D}_\mathbf{a}$ and $\mathbf{D}_\mathbf{c}$ are diagonal matrices with $\mathbf{a}$ and $\mathbf{c}$ on the main diagonal, respectively. This formulation can be applied to the whole signal $\mathit{f}$ and its corresponding confidence $\mathit{c}$ in a convolution-like structure.

\subsection{The Na\"ive Basis}
The most basic choice for the basis is a constant function, and it is denoted as the na\"ive basis. In this case, the applicability acts as a convolution filter. \textcolor{black}{This choice of basis $\mathbf{B}=\mathbf{1}$ simplifies (\ref{eq:3}) to}:
\begin{equation}\label{eq:6}
\begin{split}
\hat{\mathit{r}} &= (\mathbf{1}^* \mathbf{D}_\mathbf{a} \mathbf{D}_\mathbf{c} \mathbf{1})^{-1} \mathbf{1}^* \mathbf{D}_\mathbf{a} \mathbf{D}_\mathbf{c} \mathbf{f} \\
&= \dfrac{\mathbf{a} \cdot (\mathbf{c} \odot \mathbf{f})}{\mathbf{a} \cdot \mathbf{c}} \enspace,
\end{split}
\end{equation}
where $\odot$ is the Hadamard product and $\cdot$ is the scalar product. Note that the matrix multiplication has been replaced with point-wise operations since $\mathbf{D}_\mathbf{a}$ and $\mathbf{D}_\mathbf{c}$ are diagonal matrices. This can be formulated for the whole signal $\mathit{f}$ as a convolution operation as follows: 
\begin{equation} \label{eq:7}
\mathit{\hat{r}}[k] = \dfrac{\sum_i^n \ \mathit{a}[i] \  \mathit{f}[k-i] \ \mathit{c}[k-i]}{\sum_i^n \  \mathit{a}[i] \ \mathit{c}[k-i]} \enspace .
\end{equation}

\textcolor{black}{This formulation allows for densifying a sparse depth field $\mathit{f}$ given a point-wise confidence $\mathit{c}$ for each sample point in the depth field, where zero indicates a missing sample. With a proper choice of the applicability function $\mathit{a}$, a dense depth field is obtained.}

\subsection{The Applicability Function}\label{sec:app}
The applicability function $\mathbf{a}$ is required to be non-negative and it acts as a windowing function for the basis, \textit{e.g.} giving more importance to the central part of the signal over the vicinity. The choice of the applicability depends on the application and the characteristics of the signal. For example, for orientation estimation, it is desired that the applicability is isotropic \cite{farneback:phd_thesis}. On the other side, image inpainting requires the applicability to be anisotropic depending on the local structure of the image \cite{pham2003normalized}. The handcrafting of the applicability function is not within the scope of this paper as we aim to learn it.

\subsection{Propagating Confidences}\label{sec:conf}
A core advantage of normalized convolution is the separation between the signal and the confidence allowing confidence adaptive processing and determination of output confidence. The output confidence reflects the density of the input confidence as well as the coherence of the data under the chosen basis. Westelius \cite{Westelius302463} proposed a measure for the output confidence defined as:
\begin{equation}\label{eq:4}
\mathbf{c}_\mathrm{out} = \Big( \frac{\det \ \mathbf{G} }{\det \ \mathbf{G}_0 } \Big)^{\frac{1}{m}} \enspace,
\end{equation}
where $\mathbf{G} = \mathbf{B}^* \mathbf{D}_\mathbf{a} \mathbf{D}_\mathbf{c} \mathbf{B}$ and $\mathbf{G}_0 = \mathbf{B}^* \mathbf{D}_\mathbf{a}  \mathbf{B}$. This corresponds to a geometric ratio between the Grammians of the basis $\mathbf{B}$ in case of partial and full confidence. Similarly, Karlholm \cite{Karlholm302807} proposed another measure defined as:
\begin{equation}\label{eq:5}
\mathbf{c}_\mathrm{out} = \frac{1}{\parallel \mathbf{G}^{-1} \parallel_2 \ \parallel \mathbf{G}_0 \parallel_2}\enspace .
\end{equation}
These two measures were shown to perform well in case of the polynomial and exponential basis \cite{Westelius302463, Karlholm302807}.

\section{Unguided Normalized CNNs} \label{sec:ncnn}

Based on our prior work \cite{bmvc}, CNNs can be used to learn the optimal applicability function in case of the na\"ive basis. 

\subsection{Training the Applicability}
\begin{figure}[!t]
\centering
\subfloat[]{
\includegraphics[width=0.45\columnwidth]{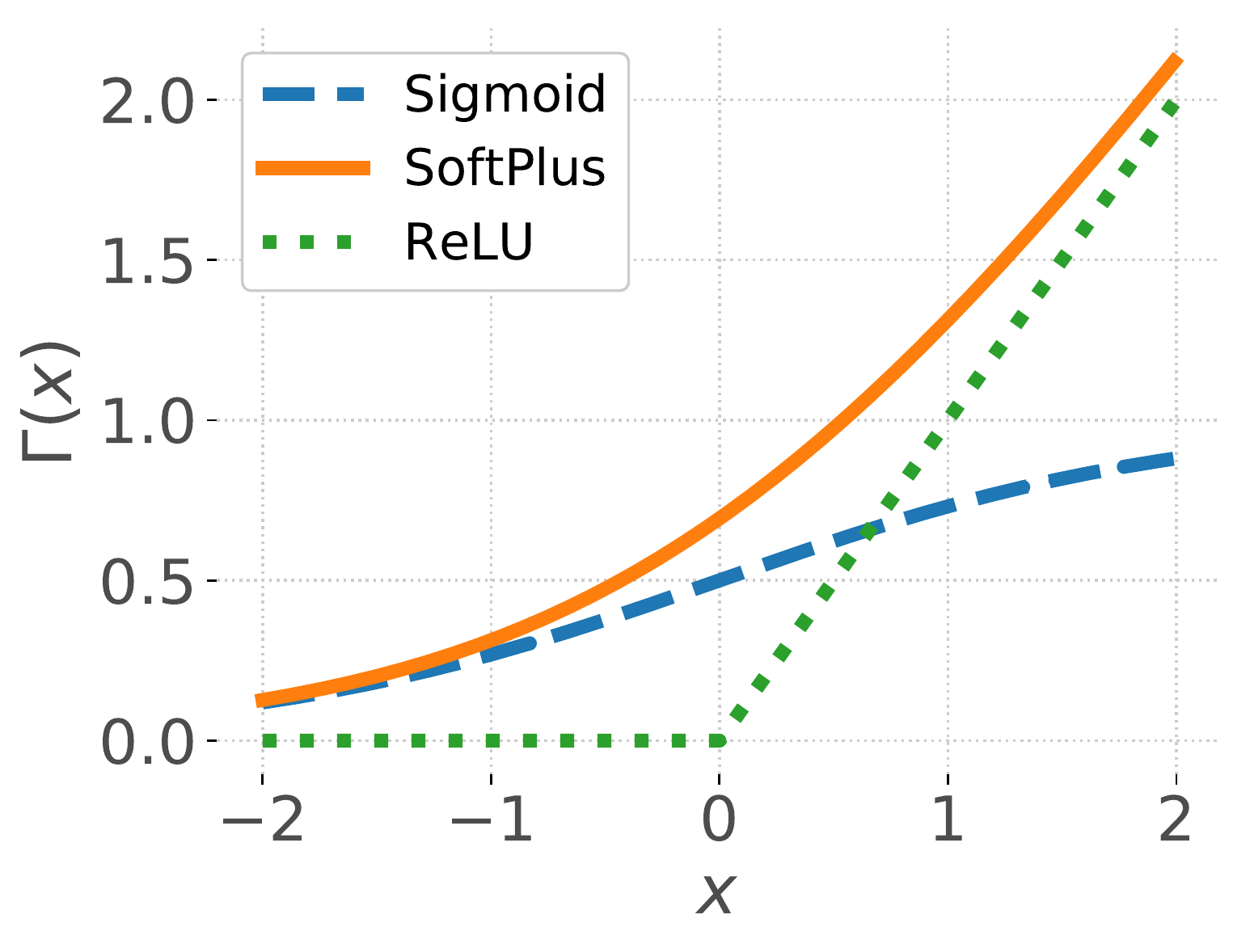}
\label{fig:nn_fn1}
}%
\hfil
\subfloat[]{
\includegraphics[width=0.50\columnwidth]{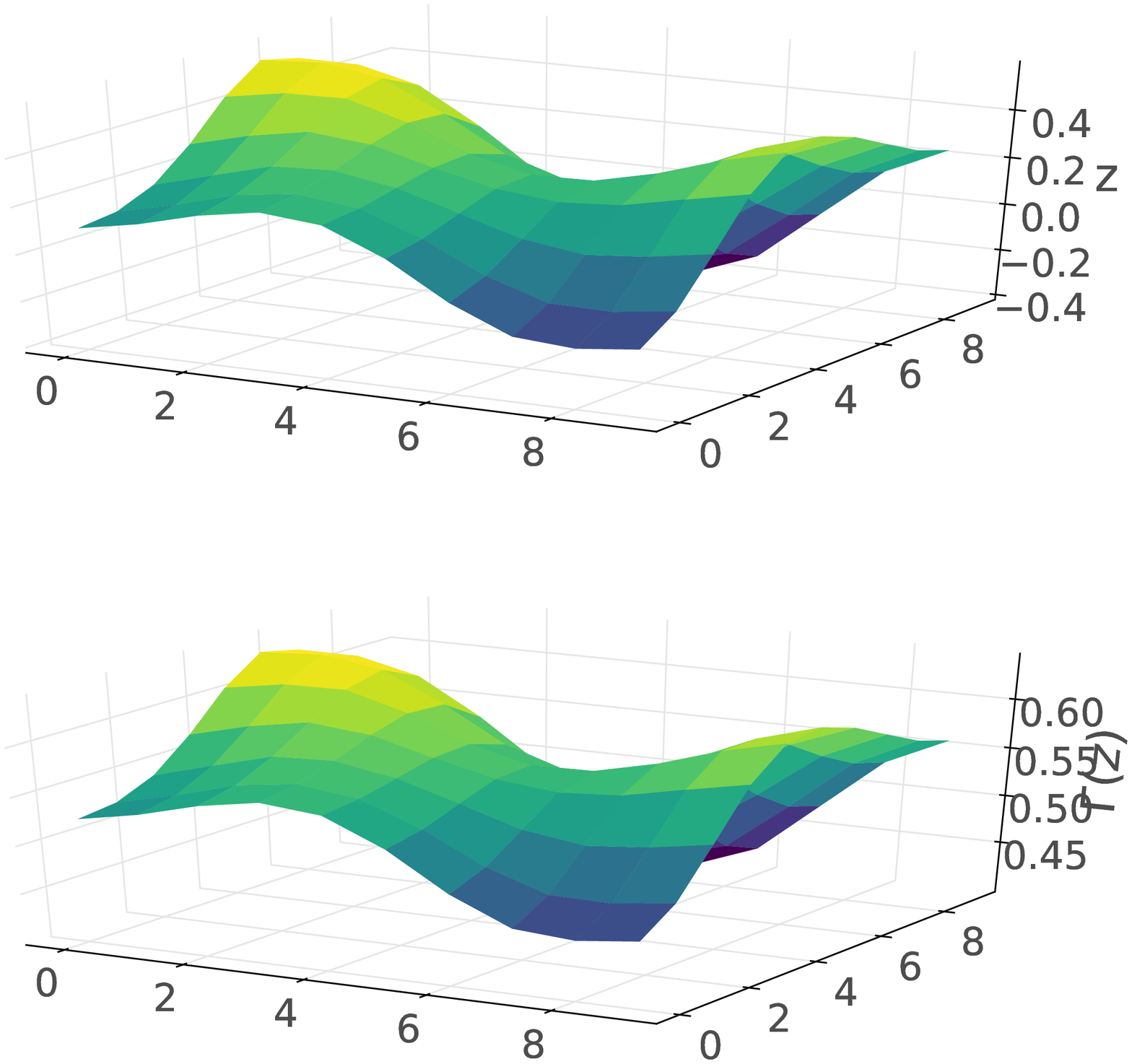}
\label{fig:nn_fn2}
}
\caption{\textbf{(a)} Examples for differentiable functions with non-negative co-domain, \textbf{(b)} Applying the SoftPlus function to a 2D surface preserves the surface trend. }
\label{fig:nn_fn}
\end{figure}

As explained in section \ref{sec:app}, the applicability is a windowing function and it needs to be non-negative. This is enforced in CNN frameworks by applying a suitable differentiable function with non-negative co-domain acting on the convolution kernels prior to the forward pass. During backpropagation, the weights will be differentiated with respect to this function using the chain rule. Examples for differentiable functions with non-negative co-domain are shown in Figure \ref{fig:nn_fn1}. Figure \ref{fig:nn_fn2} shows how the SoftPlus function, $\Gamma(z) = \log(1+\exp(z))$, translates the co-domain of a 2D surface to be non-negative while preserving the surface trend. 

Given a function $\Gamma(\cdot)$ with a non-negative co-domain, the gradients of the weight for the $l^\mathrm{th}$ convolution layer are obtained as:
\begin{equation}\label{eq:8}
\frac{\partial \mathbf{E}}{\partial \mathbf{W}^l_{m,n}} = \sum_{i,j} \frac{\partial \mathbf{E}}{\partial \mathbf{Z}^l_{i,j}} \cdot \frac{\partial \mathbf{Z}^l_{i,j}}{\partial \ \Gamma (\mathbf{W}^l_{m,n})} \cdot \frac{\partial \ \Gamma (\mathbf{W}^l_{m,n})}{\partial \mathbf{W}^l_{m,n}}\enspace,
\end{equation}
where $\mathbf{E}$ is the loss between the network output and the ground truth, and  $\mathbf{Z}^l_{i,j}$ is the output of the $\emph{l}^\mathrm{th}$ layer at locations $i,j$ depending on the weight elements $\mathbf{W}^l_{m,n}$. Accordingly, the forward pass for normalized convolution is defined as:
\begin{equation}\label{eq:9}
\mathbf{Z}^l_{i,j} =  \frac{\sum_{m,n} \ \mathbf{Z}^{l-1}_{i+m,j+n} \mathbf{C}^{l-1}_{i+m,j+n} \ \Gamma (\mathbf{W}^l_{m,n})}{ \sum_{m,n} \ \mathbf{C}^{l-1}_{i+m,j+n} \ \Gamma(\mathbf{W}^l_{m,n}) + \epsilon} + \mathbf{b}^l \enspace,
\end{equation}
where $\mathbf{C}^{l-1}$ is the output confidence from the previous layer, $\mathbf{W}^l_{m,n}$ is the applicability in this context, $\mathbf{b}^l$ is the bias and $\epsilon$ is a constant to prevent division by zero. Note that this is formally a correlation, as it is a common notation in CNNs.

\subsection{Propagating the Confidence}
The confidence output measures described in section \ref{sec:conf} have been shown to give reasonable results in case of non-na\"ive basis \cite{Karlholm302807 ,Westelius302463}. In our earlier work \cite{bmvc}, we proposed a confidence output measure for the na\"ive basis case. The proposed measure is derived from (\ref{eq:4}) and can utilize the already computed terms in the forward path. The measure is defined as:
\begin{equation}\label{eq:10}
\mathbf{C}^l_{i,j} = \frac{ \sum_{m,n} \mathbf{C}^{l-1}_{i+m,j+n} \ \Gamma( \mathbf{W}^l_{m,n}) + \epsilon}{\sum_{m,n} \ \Gamma(\mathbf{W}^l_{m,n})}\enspace.
\end{equation}
This measure allows propagating confidence between CNN layers without facing the problem of "validity masks saturation" as described in \cite{valeo}, which affects several methods in the literature \cite{Liu2018, Uhrig2017, ncnn}.

\subsection{The Normalized CNN Layer}
\begin{figure}[t!]
\centering
\includegraphics[width=\columnwidth]{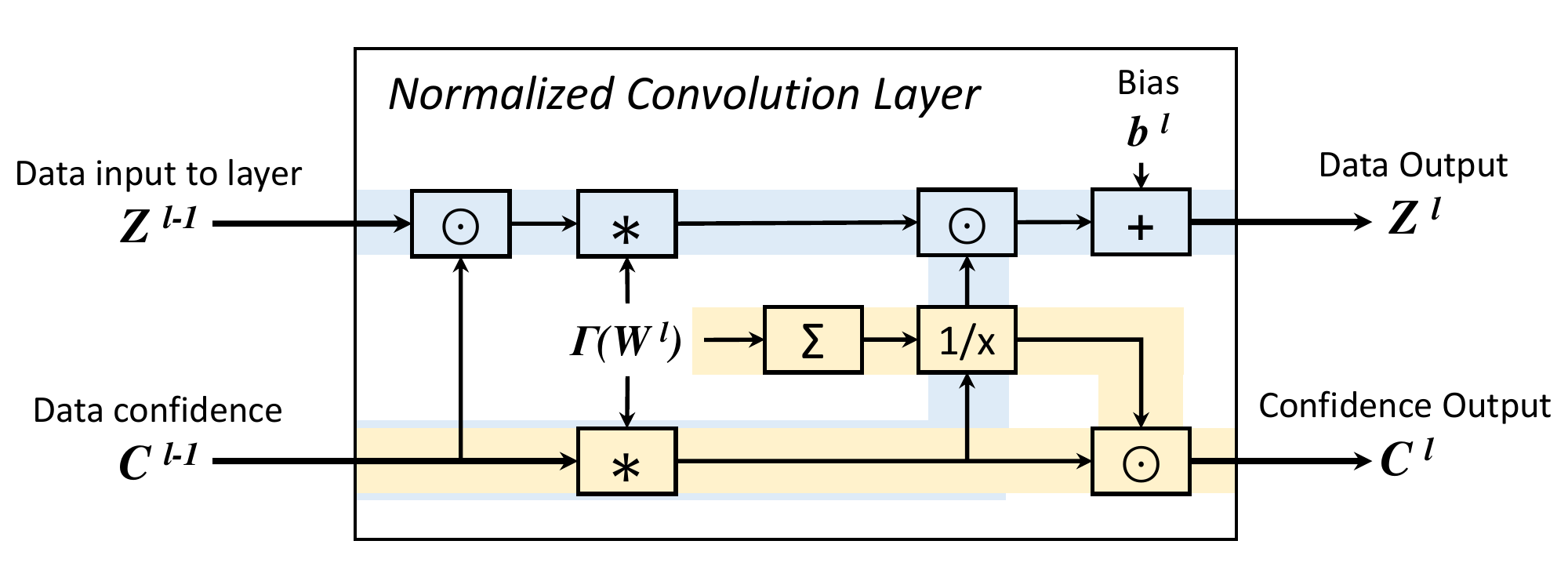}
\caption{An illustration of the Normalized Convolution layer that takes in two inputs: data and confidence. The Normalized Convolution layer outputs a data term and a confidence term. Convolution is denoted as $\ast$, the Hadamard product (point-wise) as $\odot$, summation as $\Sigma$, and point-wise inverse as $1 / \text{x}$.}
\label{fig:nconv_layer}
\end{figure}

The standard convolution layer in CNN frameworks can be replaced by a normalized convolution layer with minor modifications. First, the layer  takes in two inputs simultaneously, the data and its confidence. The forward pass is then modified according to (\ref{eq:9}) and the back-propagation is modified to include a derivative term for the non-negativity enforcement function $\Gamma(\cdot)$ as described in (\ref{eq:8}). To propagate the confidence to consecutive layers, the already-calculated denominator term in (\ref{eq:9}) is normalized by the sum of the filter elements as shown in (\ref{eq:10}). An illustration of the Normalized CNN layer is shown in Figure \ref{fig:nconv_layer}.

\begin{figure*}
\centering
\includegraphics[width=\textwidth]{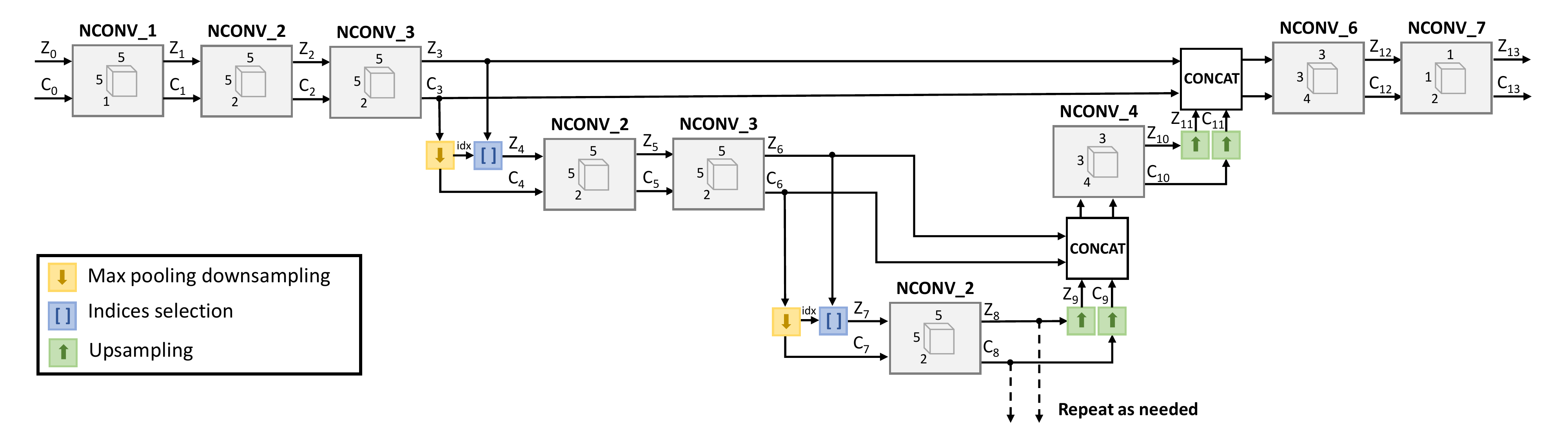}
\caption{Our proposed multi-scale architecture for the task of unguided scene depth completion that utilizes normalized convolution layers. Downsampling is performed using max pooling on confidence maps and the indices of the pooled pixels are used to select the pixels with highest confidences from the feature maps. Different scales are fused by upsampling the coarser scale and concatenating it with the finer scale. A normalized convolution layer is then used to fuse the feature maps based on the confidence information. Finally, a 1 $\times$ 1 normalized convolution layer is used to merge different channels into one channel and produce a dense output and an output confidence map. }
\label{fig:ncnn}
\end{figure*}

\subsection{The Loss Function}
In networks that perform pixel-wise tasks such as inpainting, upsampling, or segmentation, it is very common to use the \emph{L1} or the \emph{L2} norm. However, the former ignores outliers and focuses on the global level, while the latter focuses on local regions that have outliers. ِ A good compromise is the Huber norm \cite{huber1964robust}, which is defined as:
\begin{equation}\label{eq:11}
\|z- t\|_\mathrm{H} =
\begin{cases} 
\frac{1}{2}(z-t)^2, & |z-t| < \delta \\ 
\delta |z-t|-\frac{1}{2} \delta^2, & \text{otherwise}
\end{cases}
\end{equation}
The Huber norm corresponds to the \emph{L2} norm if the error is less than $\delta$ and to the \emph{L1} norm otherwise. Usually, the value of $\delta$  is set to $1$ within CNN frameworks and referred to as the \emph{Smooth L1} loss.

In networks with normalized convolution layers, it is desirable to minimize the data error and maximize the output confidence at the same time. Thus, a loss function that simultaneously achieves both objectives is desired. Assume a data error term using the Huber norm:
\begin{equation}\label{eq:12}
 \mathbf{E}_{i,j} = \|\mathbf{Z}^L_{i,j} - \mathbf{T}_{i,j} \|_\mathrm{H} \enspace, 
\end{equation} 
where $\mathbf{Z}^L_{i,j}$ is the data output from the last layer $L$ and $\mathbf{T}_{i,j}$ is the data ground truth. This is complemented with a term to maximize the confidence and the total loss $\tilde{\mathbf{E}}$ becomes:
\begin{equation}\label{eq:13}
\tilde{\mathbf{E}}_{i,j} = \mathbf{E}_{i,j} - \frac{1}{p} \left[ \mathbf{C}^L_{i,j} - \mathbf{E}_{i,j} \mathbf{C}^L_{i,j} \right] \enspace,
\end{equation}
where $\mathbf{C}^L_{i,j}$ is the output confidence and $p$ is the epoch number. The confidence term is decaying by dividing it by the epoch number $p$ to prevent it from dominating the loss when the data error term starts to converge.

\subsection{Unguided Normalized CNN Architecture}
In \cite{bmvc}, we proposed a hierarchical multi-scale architecture inspired by the U-Net \cite{Ronneberger2015}, which shares weights between different scales. The architecture acts as a generic estimator for different scales and gives a good approximation for the dense output at a very low computation cost. An illustration of the architecture is shown in Figure \ref{fig:ncnn}. At the first scale, a normalized convolution layer takes in the sparse input as well as a confidence map. Afterwards, two normalized convolution layers are applied followed by downsampling. The downsamping process is performed by applying a max pooling operator on the output confidence from the last normalized layer while keeping the indices of the pooled values as in unpooling operations \cite{zeiler2014visualizing}. These indices are then used to select the values from the features maps that have the highest confidences. This enables propagating the most confident data to the subsequent scale. To maintain the absolute levels of confidences after downsampling, we divide the downsampled confidences by the Jacobian of the scaling.

The aforementioned pipeline is repeated as required depending on the sparsity level of the data. In order to fuse different scales, the output and the corresponding confidence from the last normalized convolution layers are upsampled using nearest-neighbor interpolation and concatenated with the corresponding scale through a skip connection. After each concatenation, a new normalized convolution layer is utilized to fuse data from the two scales based on their confidences. Finally, a $1\times1$ normalized convolution layer is used to fuse different channels into one channel that corresponds to the dense output. In addition to the dense output, a confidence output is also available that holds information about the confidence distribution of the output. The output confidence can be useful for safety application or for subsequent stages in the pipeline.

\section{Guided Normalized CNNs} \label{sec:gncnn}

\begin{figure}[t]
\centering
\subfloat[Input]{\includegraphics[width=0.4\columnwidth]{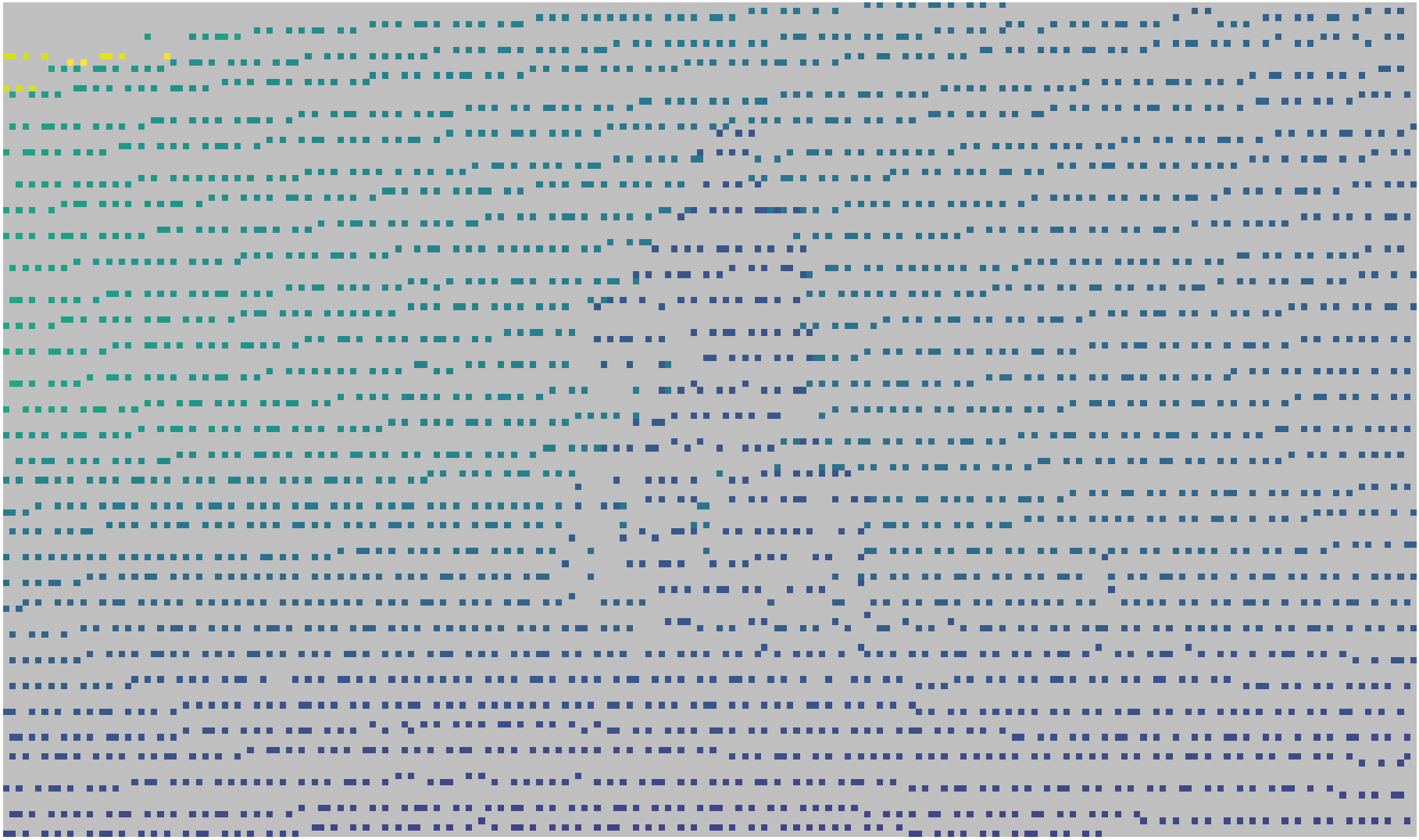}%
\label{fig:ncnn_inp}}
\hfil
\subfloat[Groundtruth]{\includegraphics[width=0.4\columnwidth]{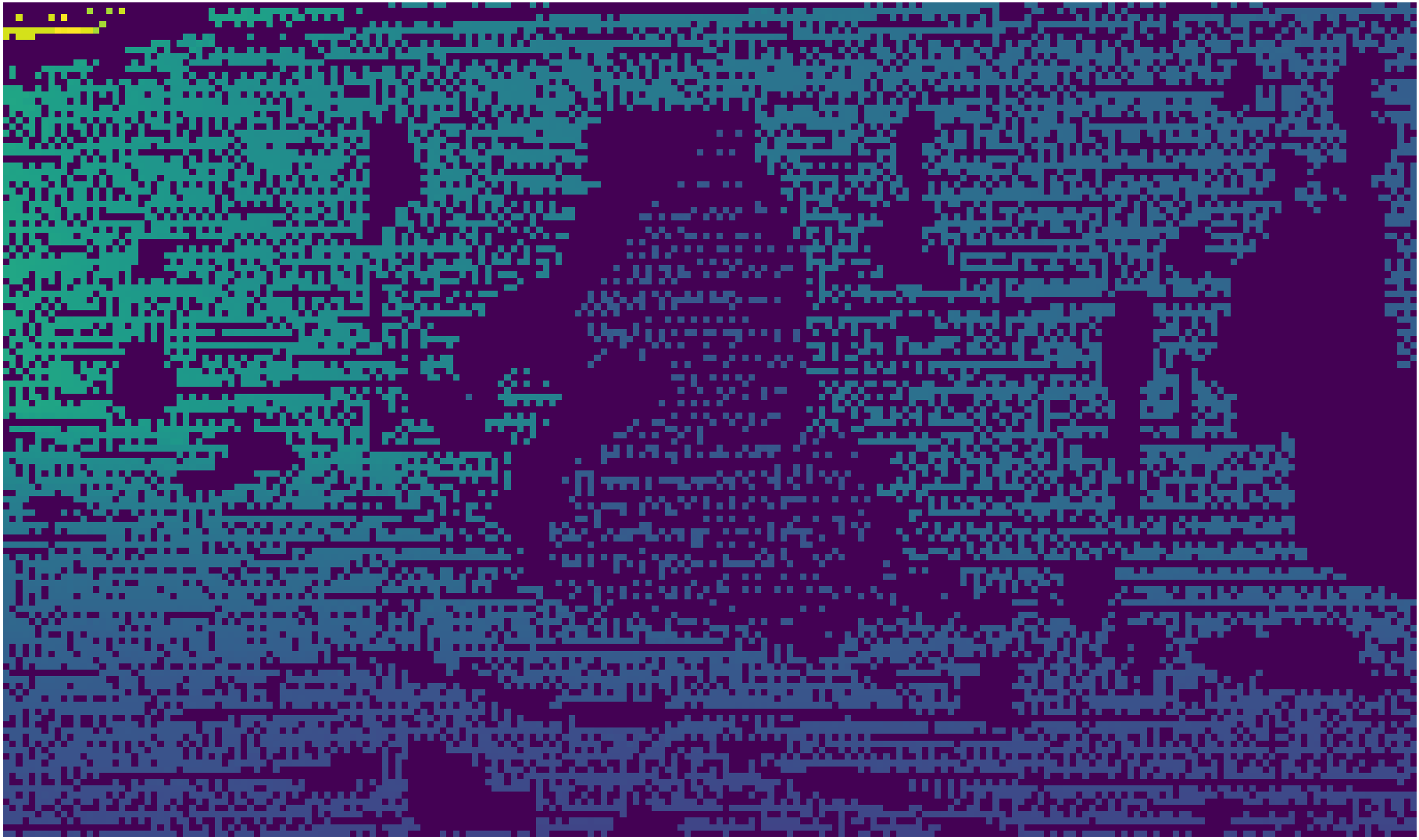}%
\label{fig:ncnn_gt}}
\vfil
\subfloat[RGB Image]{\includegraphics[width=0.4 \columnwidth]{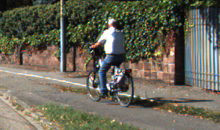}%
\label{fig:ncnn_rgb}}
\hfil
\subfloat[Absolute Error]{\includegraphics[width=0.4 \columnwidth]{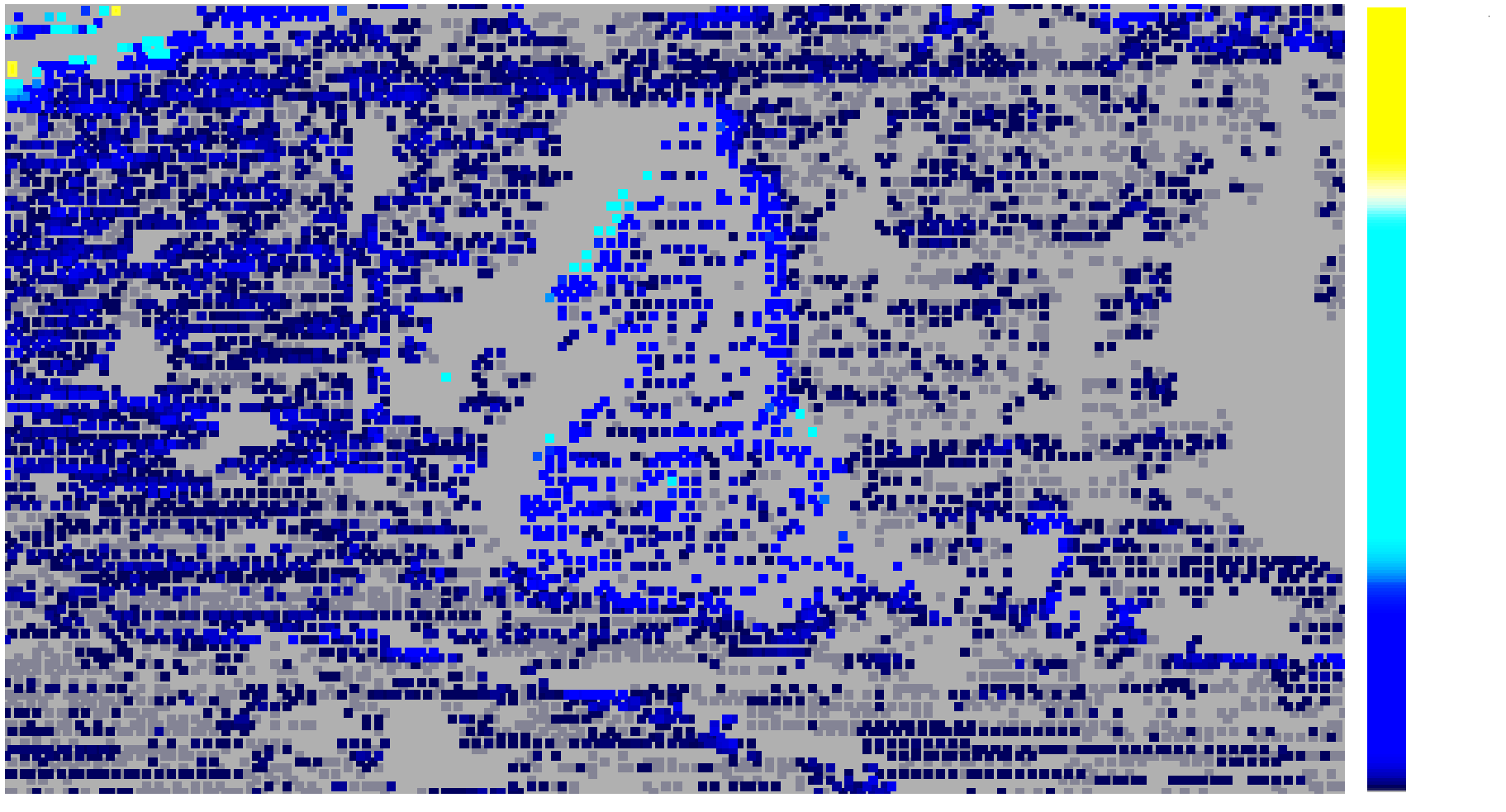}
\label{fig:ncnn_error1}}
\caption{An example for the spatial error distribution of the output from our unguided normalized convolution network on the task of depth completion. \textbf{(d)} shows that the error is distributed around edges. }
\label{fig:ncnn_error}
\end{figure}

In this section, we extend the unguided normalized convolution architecture described in section \ref{sec:ncnn} with RGB and output confidence guidance. The unguided architecture acts as a generic estimator for different scales that is learned from the data. However, this generic estimator shows weaknesses at local regions with discontinuities such as edges and rough surfaces. Figure \ref{fig:ncnn_error} shows an example of the spatial error distribution for the output from the unguided normalized convolution network on the task of depth completion. It is clear that regions with edges have larger errors than flat regions. Therefore, auxiliary data such as RGB images and surface normals can be used to alleviate this problem by providing contextual information to the network.

\subsection{RGB Image Fusion}
RGB images can be very useful in guiding the network since convolution layers typically act as feature extractors. These features are usually edges, corners, color, or texture. Providing this information to the network was demonstrated to improve the results \cite{mit,valeo,hms}, especially across edges and in rough surfaces. Therefore, we incorporate RGB information into our network to handle discontinuities at edges.

\subsection{The Output Confidence Fusion}

\begin{figure}
\begin{tabular}{ll}
& \multirow{2}{*}{\includegraphics[width=0.56 \columnwidth]{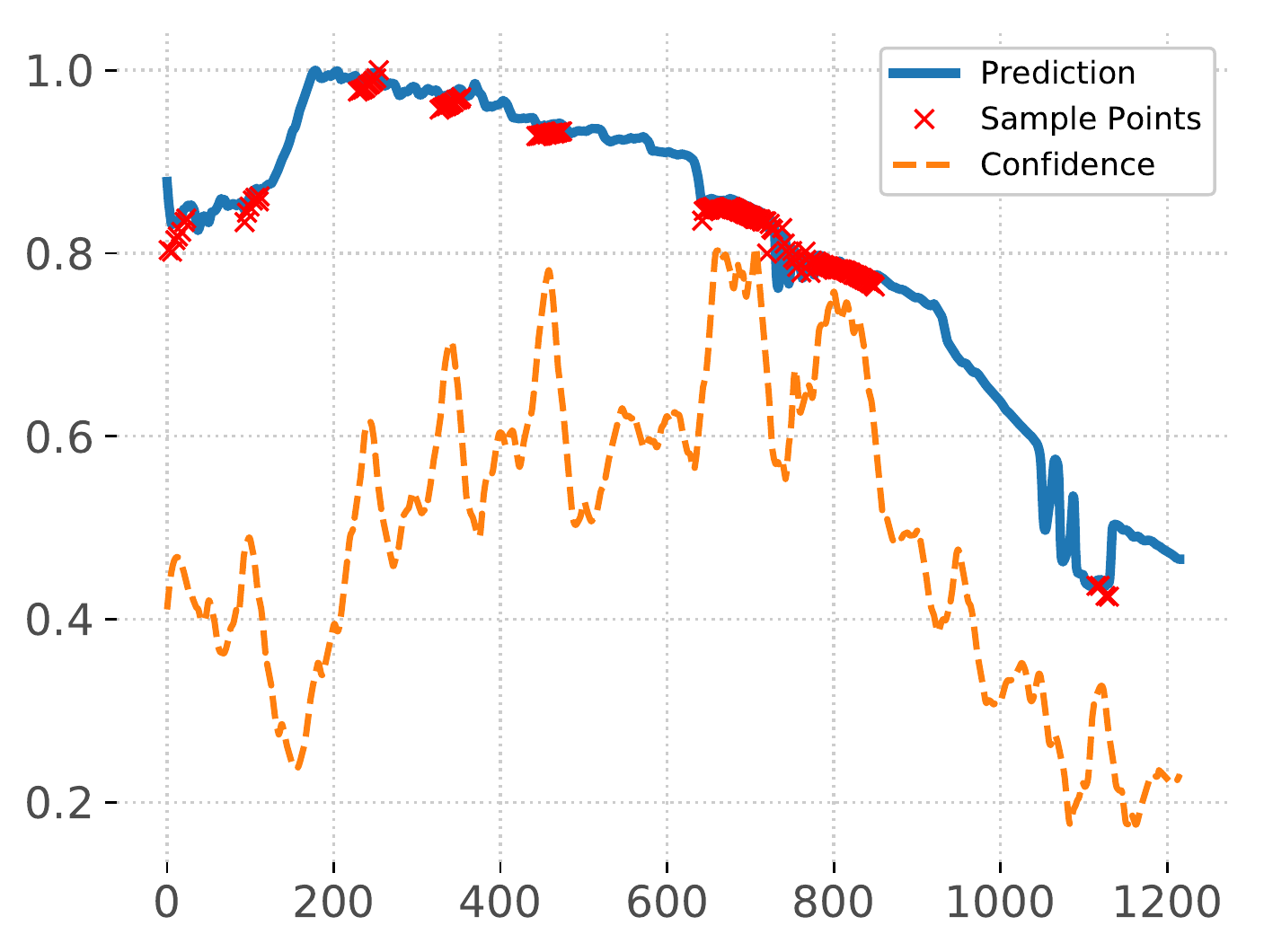}}  \\
   \multicolumn{1}{l}{\includegraphics[width=0.44\columnwidth]{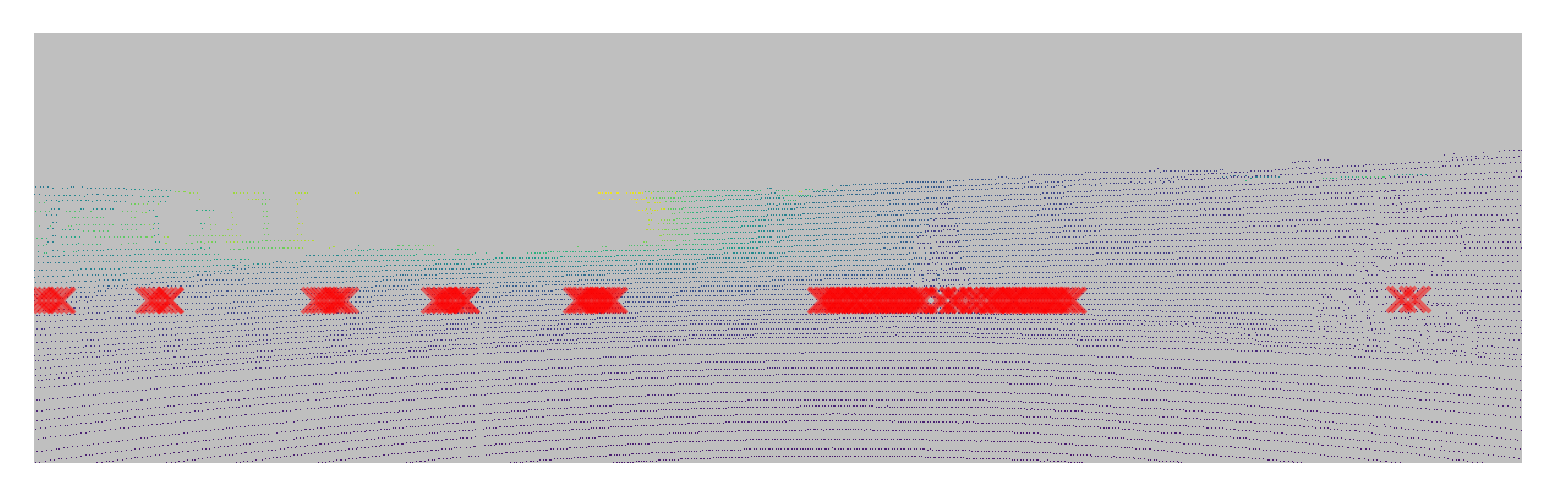}} \\
  \multicolumn{1}{l}{\includegraphics[width=0.44\columnwidth]{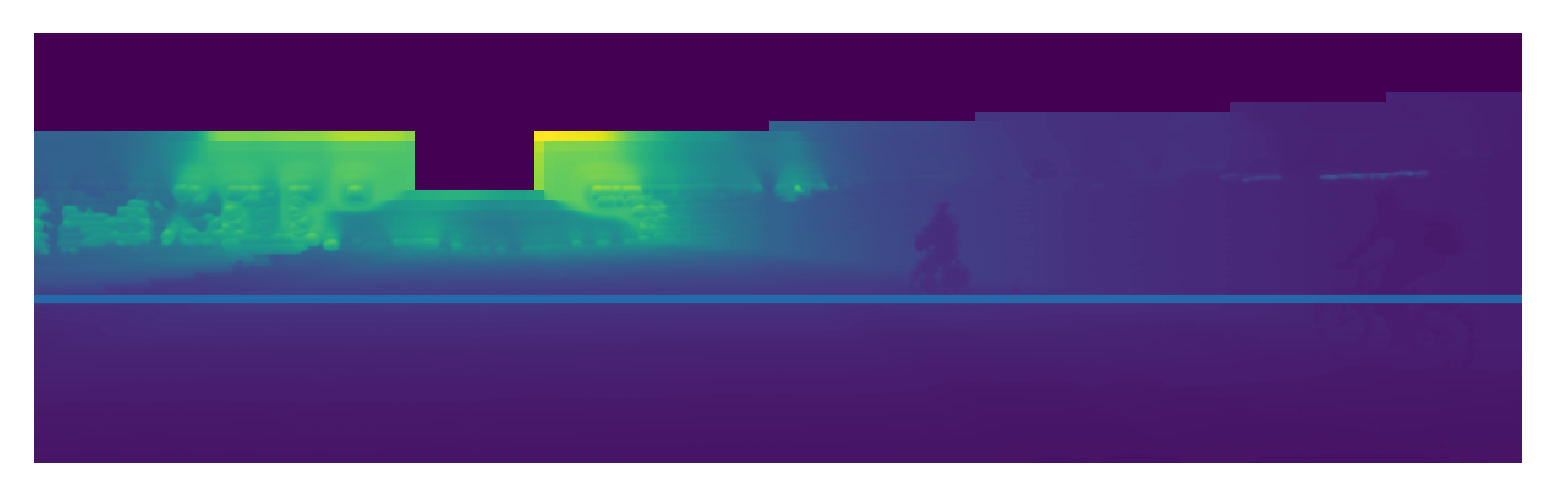}}  \\
  \multicolumn{1}{l}{\includegraphics[width=0.44\columnwidth]{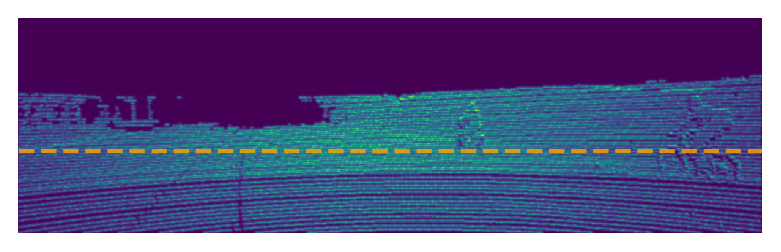}} 
\end{tabular}
\caption{An example of the output confidence from our unguided normalized convolution network on the task of depth completion. Images on the left are from top-to-bottom: sparse input, the dense output from the unguided normalized convolution network and the output confidence. The plot on the right shows the corresponding values for row 217. The red crosses are the sample points from the sparse input, the blue curve is the dense prediction and the orange curve is the output confidence (smoothed). It is shown that regions with high density of sample points tend to have a higher confidence. Note that all values are normalized to [0;1]. }
\label{fig:conf_analysis}
\end{figure}

The RGB data is fused with a new form of auxiliary data, which is the output confidence from the unguided normalized convolution network. This output confidence holds useful information about the reliability of each pixel in the image. For example, regions in the sparse input that have a high density of sample points should have a higher confidence in the output. Figure \ref{fig:conf_analysis} illustrates how the output confidence from our unguided network (the dashed orange curve) is correlated with the density of the sample points in the input (the red crosses). The figure also shows how our unguided network can find a good approximation (the blue curve) for the sparse input. Therefore, We use the output confidence as an input to our guided network to provide information about reliability of different pixels in the output from the unguided network. We will demonstrate in the experiments that the use of the output confidence improves the depth results \textcolor{black}{by} approximately $10\%$. \textcolor{black}{Further, we give statistical evidence that the output confidence correlates with the error of the prediction.}  %Both auxiliary data, the RGB information and the output confidence, are fused to guide the final layers of the network. The aim is that these produce a final, refined depth map. 

\renewcommand{\arraystretch}{0}

\begin{figure*}
\centering
\includegraphics[width=0.25\textwidth]{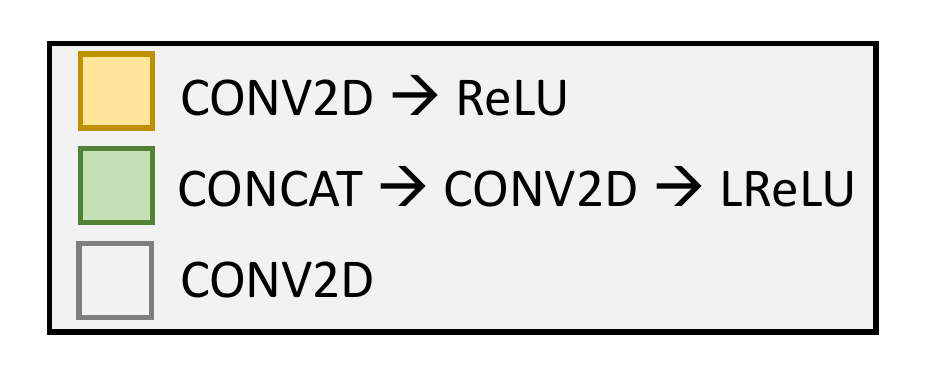}
\vfil
\subfloat[Multi-Stream (Late Fusion)]{\includegraphics[width=0.49\textwidth]{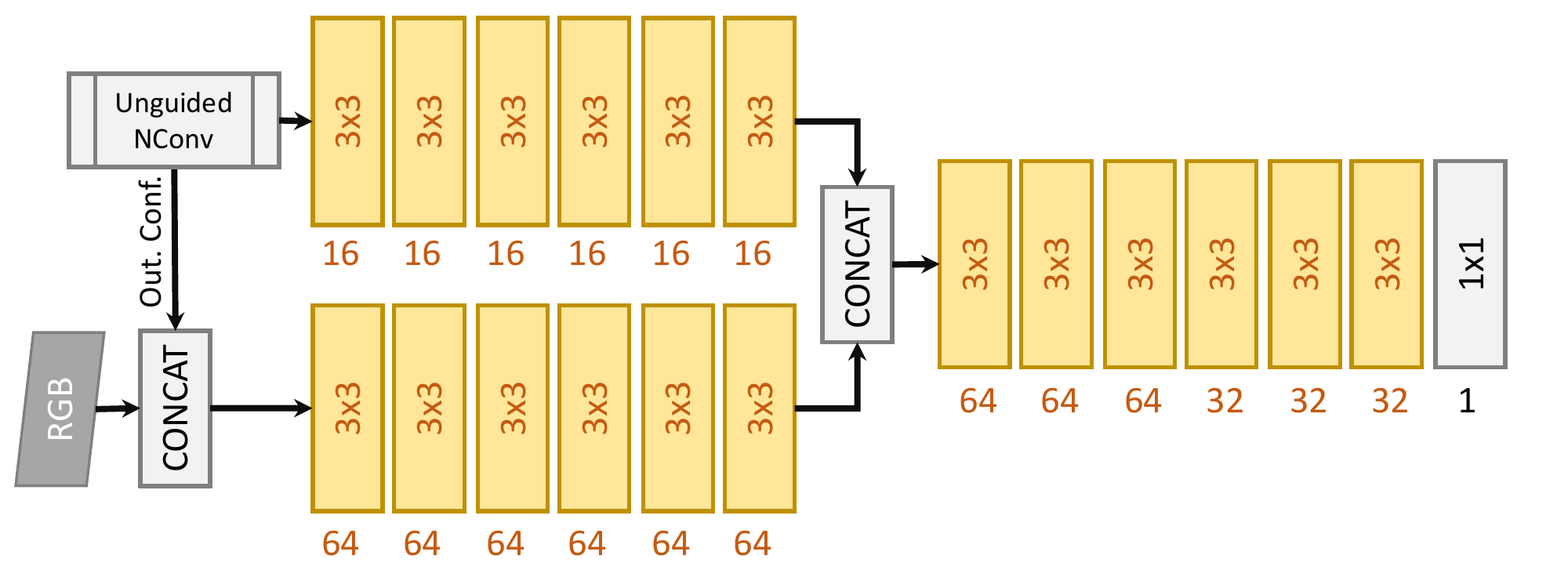}%
\label{fig:g_arch1}}
\hfil
\subfloat[Encoder-Decoder (Late Fusion)]{\includegraphics[width=0.49\textwidth]{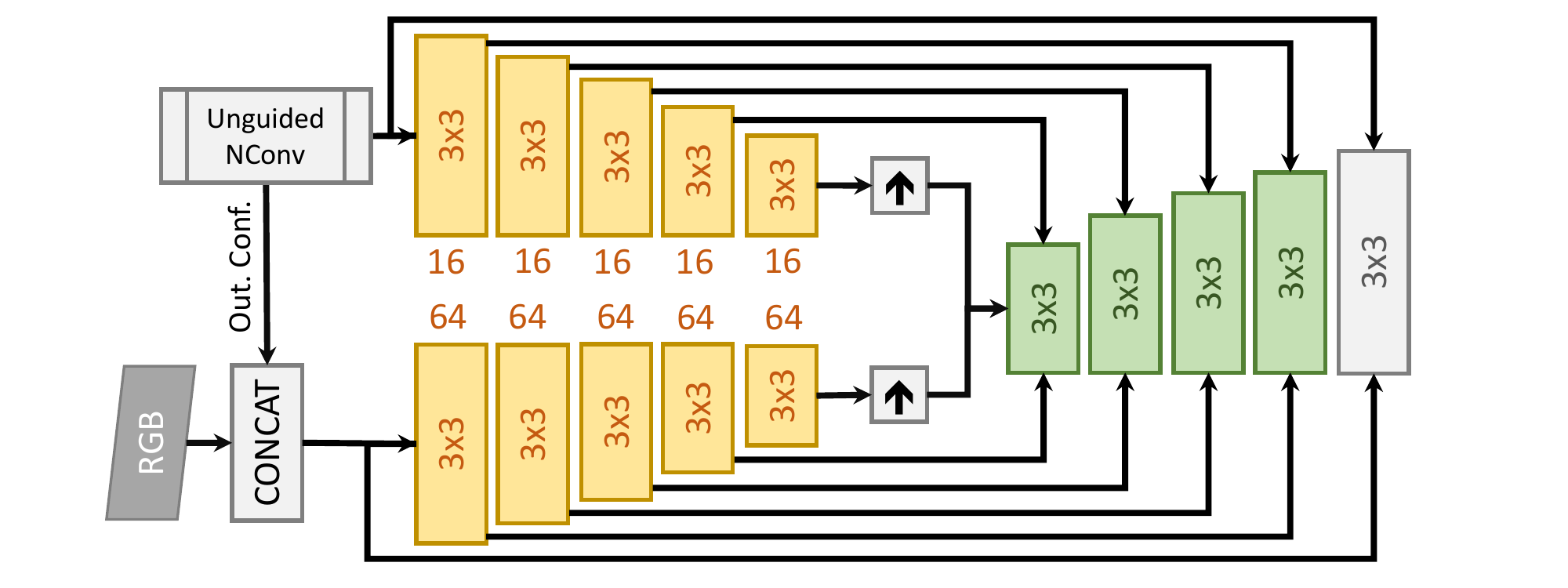}%
\label{fig:g_arch2}}
\vfil
\subfloat[Multi-Stream (Early Fusion)]{\includegraphics[width=0.49\textwidth]{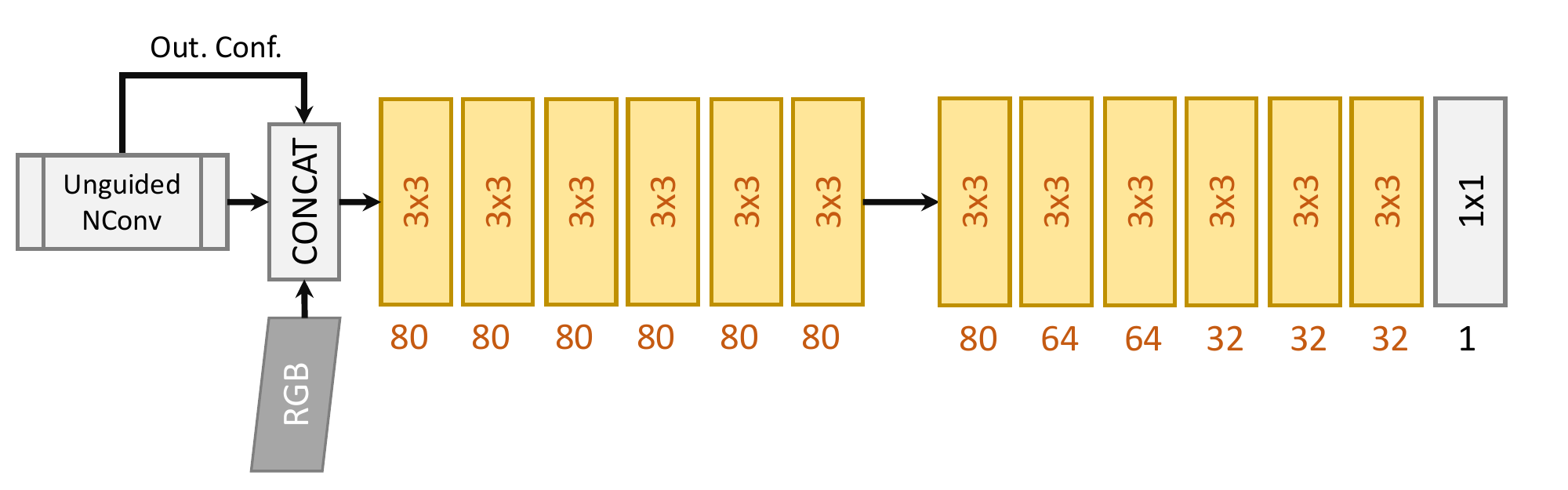}%
\label{fig:g_arch11}}
\subfloat[Encoder-Decoder (Early Fusion)]{\includegraphics[width=0.49\textwidth]{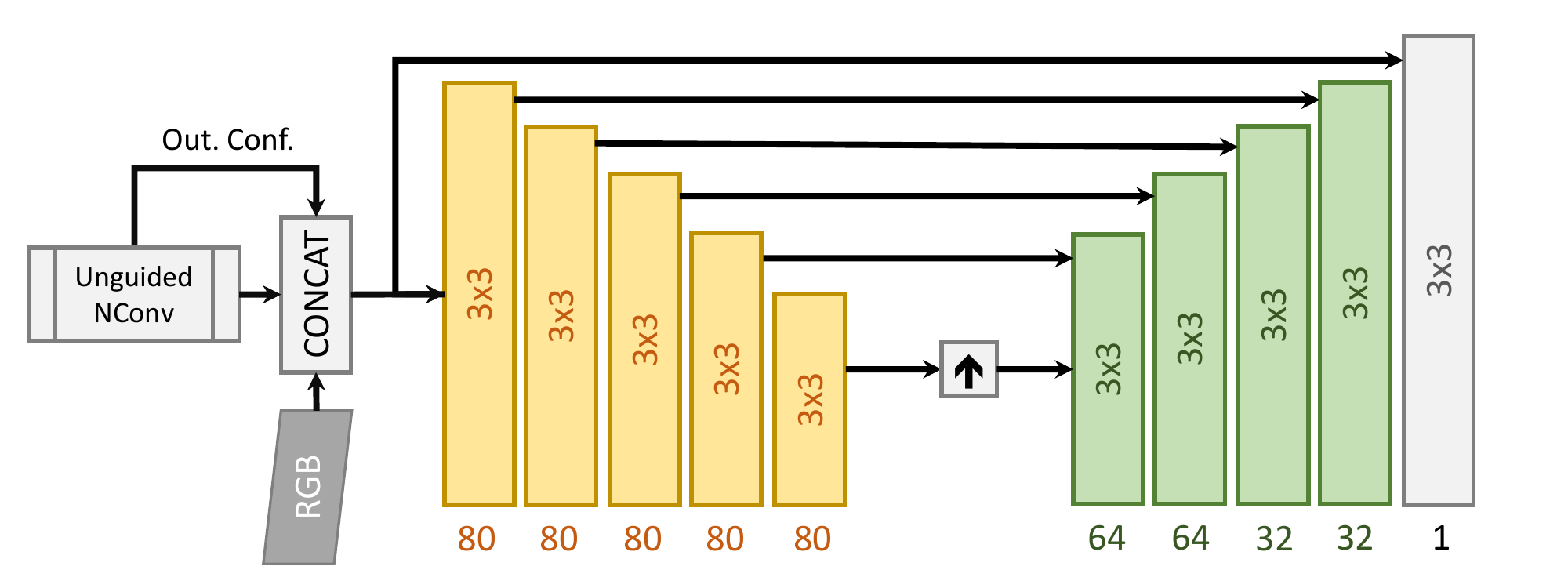}%
\label{fig:g_arch22}}
\caption{\textbf{(a)} A multi-stream architecture that contains a stream for depth and another stream for RGB+Output Confidence feature extraction. Afterwards, a fusion network combines  both streams to produce the final dense output. \textbf{(d)} A multi-scale encoder-decoder architecture where depth is fed to the unguided network followed by an encoder and output confidence and RGB image are concatenated then fed to a similar encoder. Both streams have skip-connection to the decoder between the corresponding scales. \textbf{(c)} is similar to \textbf{(a)}, but with early fusion and \textbf{(d)} is similar to \textbf{(b)} but with early fusion.}
\label{fig:g_arch}
\end{figure*}

\subsection{Network Architecture}\label{sec:g_arch}

\renewcommand{\arraystretch}{1.3}
\begin{table}[b]
\begin{center}
\begin{tabular}{ C{3.5cm} | C{2.2cm} | C{2.2cm} }
\specialrule{.2em}{.1em}{.1em}
& Early Fusion & Late Fusion \\
\hline
Multi-stream &  $\bullet$ & \cite{ncnn} \\
\hline
Encoder-decoder & \cite{Liu2018, mit} & \cite{hms,valeo}\\
\specialrule{.2em}{.1em}{.1em} 
\end{tabular}
\end{center}
\caption{A categorization of the state-of-the-art methods depending on the architecture and the fusion scheme. }
\label{tab:sch}
\end{table} 

We aim to fuse the sparse depth, the RGB image, and the output confidence to produce a dense depth map. Therefore, we look into two of the commonly used architectures from the literature on the task of guided scene depth completion. The first is a simple multi-stream network inspired by \cite{ncnn} which shares similarities with our proposed work and the second is an encoder-decoder architecture with skip-connections inspired by \cite{Ronneberger2015, Liu2018, mit}, which was demonstrated to achieve state-of-the-art results. The former employs a late-fusion scheme for combining different streams, while the latter adopts an early-fusion scheme. Table \ref{tab:sch} summarizes some methods in the literature under this categorization.

We investigate all cases from Table \ref{tab:sch} with both architectures and both fusion schemes. First, we utilize a multi-stream network with late fusion as is shown in Figure \ref{fig:g_arch1}. One stream contains our unguided network described in section \ref{sec:ncnn} followed by refinement layers and the other contains the image concatenated with the output confidence from the unguided network. Eventually, both streams are fused  by concatenation and then fed to a fusion network that produces the final output. In addition, we train the same network in an early fusion manner as illustrated in Figure \ref{fig:g_arch11}. Note that the number of channels for the depth stream was added to the RGB stream, while the number of channels for the fusion network were kept unchanged.

Secondly, we adopt a multi-scale encoder-decoder network with late fusion as shown in Figure \ref{fig:g_arch2}. One stream contains our unguided network followed by an encoder, where all convolution layers apply a stride of 2 to perform downsampling and a ReLU activation. In the other stream, both RGB image and output confidence from the unguided network are concatenated and then fed to an encoder similar to the previous one, but with a larger number of channels per layer. At the decoder, feature maps from both streams are upsampled and then concatenated with the feature maps having the same scale from the encoder. Afterwards, convolution is performed followed by Leaky ReLU activation with $\alpha=0.2$. The final layer produces the final dense output. Similarly, we also apply an early fusion scheme to this architecture by concatenating both the output and the output confidence from the unguided network with the RGB image. Then, they are fed into a similar encoder-decoder as illustrated in Figure \ref{fig:g_arch22}. In this way, we evaluate all options listed in Table \ref{tab:sch}.

For all networks described here, we aim to reduce the number of parameters for computational efficiency. Therefore, we use a fixed number of feature channels of 16 per each input channel. For example, sparse depth input has only one channel, so we use 16 features at all layers in the depth stream, while we use $16 \times 4 = 64$ for the RGB image and the output confidence. Our experiments will demonstrate how the use of the unguided normalized convolution sub-network allows achieving state-of-the-art results on the task of guided depth completion without requiring huge networks with millions of parameters as in \cite{mit,valeo,hms}.

\section{Experiments}
\label{sec:exp}
To demonstrate the capabilities of our proposed approach, we evaluate our proposed networks on the KITTI-Depth Benchmark \cite{Uhrig2017} and the NYU-Depth-v2 \cite{nyu} dataset for the task of depth completion. 

\subsection{Datasets}
\textbf{KITTI-Depth dataset} \cite{Uhrig2017} includes depth maps from projected LiDAR point clouds that were matched against the depth estimation from the stereo cameras. The depth images are highly sparse with only ~5\% of the pixels available and the rest is missing. The dataset has 86k training images, 7k validation images, and 1k test set images on the benchmark server with no access to the ground truth. The test set will be used for evaluation against other methods, while a subset of the validation set (1k images) will be used for the analysis of our own method. We also use the RGB images from the raw data of the original KITTI dataset \cite{kitti}. It is worth mentioning that the groundtruth of the KITTI-Depth dataset is incomplete since pixels that were not consistent with the groundtruth form the stereo disparity have been removed.

\textcolor{black}{\textbf{The NYU-Depth-v2 dataset} \cite{nyu} is an RGB-D dataset for indoor scenes, captured with a Microsoft Kinect. We train a model that produces a dense depth map using the RGB image and a uniformly sampled depth points. Similar to \cite{Ma2017SparseToDense, cheng2018depth}, we use the official split with roughly 48k RGB-D pairs for the training and 654 pairs for testing. To match the resolution of the RGB images and the depth maps, the RGB images of size $640\times480$ are downsampled and center-cropped to $304 \times 228$ as described in \cite{eigen2014depth}.}

\subsection{Experimental Setup}
All experiments were performed on a workstation with 6 CPU cores, 112 GB of RAM, and an NVIDIA Tesla V100 GPU with 16 GB of memory. All guided networks were trained until convergence on the full training set \textcolor{black}{with a batch size of 4 and 8 for the KITTI-Depth dataset and the NYU-Depth-v2 datasets, respectively. We use the ADAM optimizer with an initial learning rate of $10^{-4}$ and a decaying factor of 0.1 after 20 epochs for the KITTI-Depth dataset and 10 epochs for the NYU-Depth-v2 dataset}. When only the unguided normalized convolution network is trained on the KITTI-Depth dataset, we train on 10k images of the training set for 5 epochs using the ADAM optimizer with a learning rate of 0.01. \textcolor{black}{We have implemented our network using the PyTorch framework and the source code is available on Github.} \footnote{https://github.com/abdo-eldesokey/nconv}

\subsection{Evaluation Metrics}
\textcolor{black}{For the KITTI-Depth dataset}, we adopt the evaluation metrics used in the benchmark \cite{Uhrig2017}: the Mean Absolute Error (MAE) and the Root Mean Square Error (RMSE) computed on the depth values. The MAE is an unbiased error metric takes takes an average of the error across the whole image and it is defined as:
\begin{equation}
\text{\emph{MAE}}(Z, T) = \frac{1}{MN} \left[ \sum_{i=0}^{N} \sum_{j=0}^{M} | Z(i,j) - T(i,j)| \right]  \enspace ,
\end{equation}\label{eq:14}
while the RMSE penalizes outliers and it is defined as:
\begin{equation}
\text{\emph{RMSE}}(Z, T) = \frac{1}{MN} \left[ \sum_{i=0}^{N} \sum_{j=0}^{M} | Z(i,j) - T(i,j)|^2 \right]^{1/2} \enspace.
\end{equation}\label{eq:16}
\noindent Additionally, we also use iMAE and iRMSE, which are calculated on the disparity instead of the depth. \textcolor{black}{The 'i' indicates that disparity is proportional to the inverse of depth}. 

\textcolor{black}{For the NYU-Depth-v2 dataset, we compute the RMSE, the mean absolute relative error (REL), and the inliers ratio as descriped in \cite{eigen2014depth}.}

\subsection{Evaluating Guided Normalized CNNs} \label{sec:fusion}
First, we evaluate the different architectures described in section \ref{sec:g_arch} on the KITTI-Depth dataset \cite{Uhrig2017}. The first architecture is a multi-stream network with early fusion denoted as \emph{MS-Net[EF]} and its variant with late fusion \emph{MS-Net[LF]}. The second architecture is an encoder-decoder architecture with early-fusion denoted as \emph{EncDec-Net[EF]} and its variant with late fusion \emph{EncDec-Net[LF]}. For a fair comparison and to neutralize any influence from inefficient gradients, we train the unguided network separately using our proposed loss in (\ref{eq:13}) and then attach it to the guided networks in comparison while freezing its weights. 

All networks are trained using the Huber norm loss described in (\ref{eq:11}). Table \ref{tab:fusion} shows that the multi-stream network with late fusion, \emph{MS-Net[LF]}, outperforms all the other networks with respect to all evaluation metrics. The multi-stream network with early fusion, \emph{MS-Net[EF]}, achieves similar results with respect to MAE and iMAE, but the RMSE and iRMSE are slightly higher. For the encoder-decoder networks, \emph{EncDec-Net[EF]} with early fusion achieves better results than the network with late fusion contrarily to the multi-stream network.

Next, we compare our best performing architectures using multi-stream, \emph{MS-Net[LF]}, and encoder-decoder, \emph{EncDec-Net[EF]}, against state-of-the-art methods on the KITTI-Depth and NYU-Depth-v2 datasets.

\renewcommand{\arraystretch}{1.3}
\begin{table}
\begin{center}
\begin{tabular}{L{3.5cm} | C{1.2cm} C{1.2cm} C{1cm} C{1cm} }
\specialrule{.2em}{.1em}{.1em} 
& MAE [mm] & RMSE [mm] & iMAE [1/km] & iRMSE [1/km] \\
\hline
MS-Net[LF] & \textbf{209.56} & \textbf{908.76} & \textbf{0.90} &\textbf{2.50} \\ % Exp_3ffull2
MS-Net[EF] &  209.75 & 932.01& 0.92 & 2.64 \\ % Exp_3ffull11
EncDec-Net[LF] &  295.92 & 1053.91 & 1.31 & 3.42 \\ % Exp_4h
EncDec-Net[EF] &  236.83 & 1007.71 & 0.99 & 2.75 \\ % Exp_4c
\specialrule{.2em}{.1em}{.1em} 
\end{tabular}
\end{center}
\caption{A quantitative comparison between different fusion schemes (described in section \ref{sec:gncnn}) on the selected \textbf{validation} set of the KITTI-Depth dataset \cite{Uhrig2017}. The different fusion schemes are \emph{MS-Net[LF]}, which is the multi-stream architecture with late fusion (Figure \ref{fig:g_arch1}), \emph{MS-Net[EF]}, which applies early fusion (Figure \ref{fig:g_arch11}), \emph{EncDec-Net [LF]}, which is the encoder-decoder architecture with late fusion (Figure \ref{fig:g_arch2}), and \emph{EncDec-Net [EF]}, which applies early fusion (Figure \ref{fig:g_arch22}). \emph{MS-Net[LF]} achieves the best results with respect to all evaluation metrics.}
\label{tab:fusion}
\end{table}

%%%%%%%%%%%%%%%%%%%%%%%%%%%%%%%%%%%%%%%%%%%%%%%%%%%%%%%%%%%%%%%

\subsection{The KITTI-Depth Dataset Comparison}
We compare \emph{MS-Net[LF]} and \emph{EncDec-Net[EF]} against all published methods that have been submitted to the KITTI-Depth benchmark \cite{Uhrig2017}. \emph{SparseConv} \cite{Uhrig2017} proposed a sparsity invariant layer that normalizes the sparse input using a binary validity mask. They also created three baselines: \emph{CNN}, which trains a simple network directly on the sparse input, \emph{CNN+mask}, which concatenates the validity mask with the sparse input and trains the same network, and \emph{NN+CNN}, which performs nearest neighbor interpolation on the sparse input and then trains a refinement network. \noindent \emph{ADNN} \cite{Chodosh2018} employed compressed sensing within CNNs to handle the sparsity in data. \emph{IP-Basic} \cite{ipbasic} applied an extensive search on variations of morphology and simple image processing techniques to interpolate the sparse input. \emph{Spade} \cite{valeo} proposed an encoder-decoder architecture with late-fusion to reconstruct a dense output from the sparse input. \emph{Sparse-to-Dense} \cite{mit} proposed a self-supervised approach to alleviate the incomplete groundtruth in the KITTI-Depth dataset. Their self-supervised approach requires a sequence of sparse depth and RGB images to reconstruct a dense depth map. Finally, \emph{HMS-Net} derived variations of the sparsity invariant layer that were used to deploy larger and more complex networks.

%\textcolor{red}{As mentioned earlier, our \emph{MS-Net[LF]} with late fusion provides the best results. We compare our \emph{MS-Net[LF]} architecture against the state-of-the-art methods. Since the KITTI-Depth benchmark ranks methods based on the RMSE error, we train the same network but with an L2-norm loss and we denote it as \emph{MS-Net[LF]-L2}. Note that \emph{NConv-CNN (d)} is our unguided normalized convolution network described in section \ref{sec:ncnn}. [REMOVE]} 

\renewcommand{\arraystretch}{1.3}
\begin{table}
\begin{center}
\begin{tabular}{L{3.7cm} | C{1.2cm} C{1.2cm} C{1cm} C{1cm} }
\specialrule{.2em}{.1em}{.1em} 
 & MAE [mm] & RMSE [mm] & iMAE [1/km] & iRMSE [1/km] \\
\hline
CNN \cite{Uhrig2017} & 620.00 & 2690.00 & - & - \\
CNN+mask \cite{Uhrig2017} &  790.00 & 1940.00 & - & - \\
SparseConv \cite{Uhrig2017} & 481.27 & 1601.33 & 1.78 & 4.94 \\
NN+CNN \cite{Uhrig2017} & 416.14 & 1419.75 & 1.29 & 3.25 \\
ADNN \cite{Chodosh2018} & 439.48 & 1325.37 & 3.19	& 59.39	\\
IP-Basic \cite{ipbasic} & 302.60 & 1288.46	& 1.29	& 3.78	\\
NConv-CNN (d) \textbf{(ours)} & 360.28 & 1268.22 & 1.52	& 4.67 \\
Spade (d) \cite{valeo} & 248.32 & 1035.29 & 0.98 & 2.60	\\
Sparse-to-Dense (d) \cite{mit} & 288.64	& 954.36 & 1.35	& 3.21	\\
HMS-Net (d) \cite{hms} & 258.48	& 937.48 & 1.14	& 2.93 \\
\hline
\textcolor{black}{EncDec-Net[EF]-L1} \textbf{(ours)} & 239.39 & 965.45 & 1.01 &2.60  \\
Spade (gd) \cite{valeo} & 234.81 & 917.64 &\textit{ 0.95} & \textbf{2.17} \\
\textcolor{black}{MS-Net[LF]-L1} (gd) \textbf{(ours)} & \textbf{207.77} & 859.22 & \textbf{0.92} &\textit{2.52} \\
HMS-Net (gd) \cite{hms} & 253.47 & 841.78 & 1.13 & 2.73	\\
\textcolor{black}{MS-Net[LF]-L2} (gd) \textbf{(ours)} & \textit{233.26}	& \textit{829.98} & 1.03	& 2.60 \\
Sparse-to-Dense (gd) \cite{mit} & 249.95	& \textbf{814.73} & 1.21	& 2.80	\\
\specialrule{.2em}{.1em}{.1em} 
\end{tabular}
\end{center}
\caption{Quantitative results for methods in comparison on the KITTI-Depth benchmark \cite{Uhrig2017}. The best method is shown in \emph{bold} and the second best is shown in \textit{italic}. The performance is shown on the \emph{test} set for which the results were submitted to the benchmark evaluation server with no access to the ground truth. For all methods, \emph{(d)} denotes that only the sparse depth input was used, while \emph{(gd)} denotes that both the sparse depth and the RGB images were used. Our method \emph{\textcolor{black}{MS-Net[LF]-L1} (gd)} outperforms all methods with respect to the MAE error with a large margin. Our method \emph{\textcolor{black}{MS-Net[LF]-L2} (gd)} trained using the L2-norm achieves second best results with respect to RMSE with a small margin to the top-performing method \emph{Sparse-to-Dense (gd)}.} 
\label{tab:1}
\end{table}

Quantitative results on the test set of the KITTI-Depth dataset \cite{Uhrig2017} using evaluation metrics described above are shown in Table \ref{tab:1}. Our method \textcolor{black}{\emph{MS-Net[LF]-L1 (gd)}} outperforms all the other methods with respect to the MAE with a large margin. When trained using the L2-norm, it was able to perform the second best with respect to RMSE with a small margin compared to \emph{Sparse-to-Dense (gd)}. However, our method has a significantly lower number of parameters ($\sim$ 355k) compared to \emph{Sparse-to-Dense (gd)} which has $\sim$ 5.5M parameters. This demonstrates that our method achieves state-of-the-art results while requiring a very small number of parameters. \textcolor{black}{On the other hand, our method \emph{EncDec-Net[EF]-L1} achieves moderate results.}  \emph{Spade (gd)} on the other hand achieves the best results with respect to iRMSE since it was trained on disparity using the L2-norm. However, our method \textcolor{black}{\emph{MS-Net[LF]-L1 (gd)}} still outperformed \emph{Spade (gd)} with respect to iMAE despite being trained on depth. 

\begin{figure}
\begin{tabular}{C{0.05\columnwidth} C{0.45\columnwidth} C{0.45\columnwidth} }
%RGB image & NConv-CNN-L2 (gd) (Ours) & Sparse-to-Dense (gd) \cite{mit} & HMS-Net (gd) \cite{hms} \\
(a) & \includegraphics[width=0.45\columnwidth]{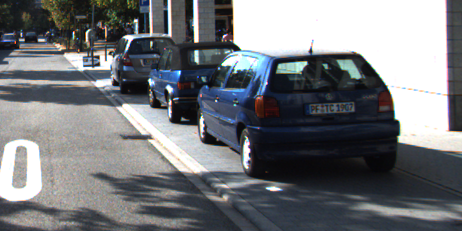} & 
\includegraphics[width=0.45\columnwidth]{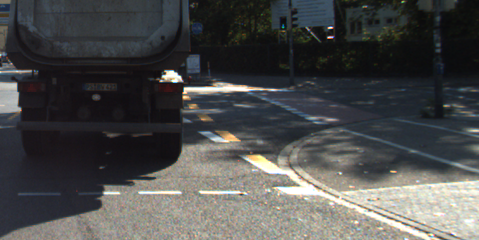} \\
(b) & \includegraphics[width=0.45\columnwidth]{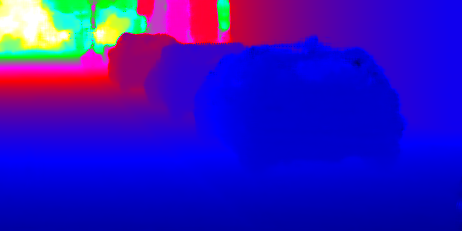} &
   \includegraphics[width=0.45\columnwidth]{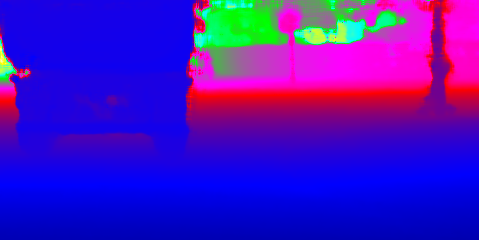} \\   
&   \includegraphics[width=0.45\columnwidth]{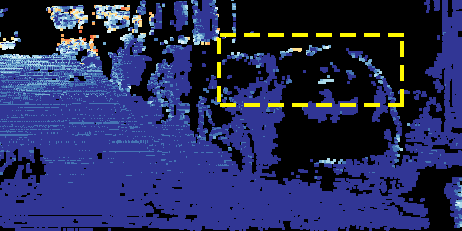} &
 \includegraphics[width=0.45\columnwidth]{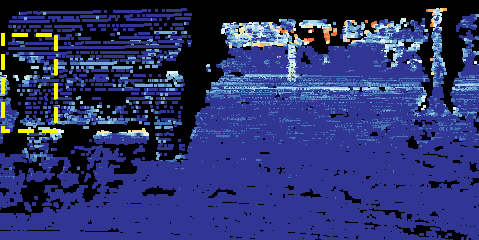}   \\
(c) & \includegraphics[width=0.45\columnwidth]{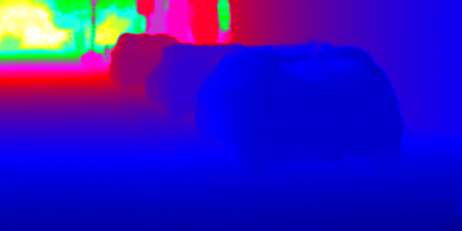} &
\includegraphics[width=0.45\columnwidth]{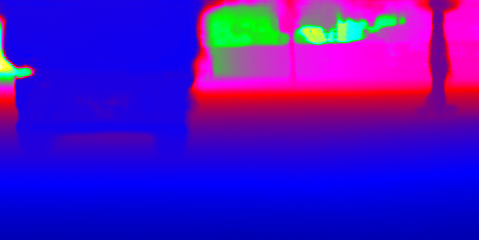} \\
& \includegraphics[width=0.45\columnwidth]{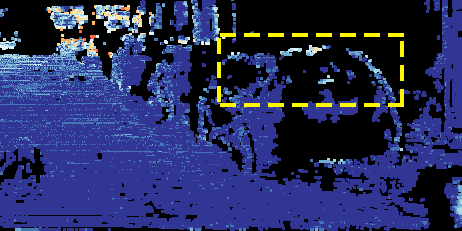} &
 \includegraphics[width=0.45\columnwidth]{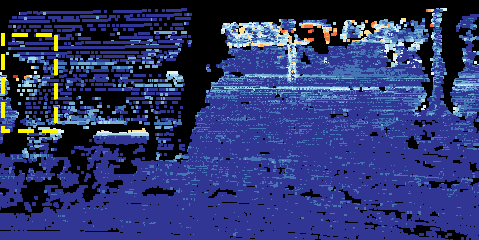} \\
(d) & \includegraphics[width=0.45\columnwidth]{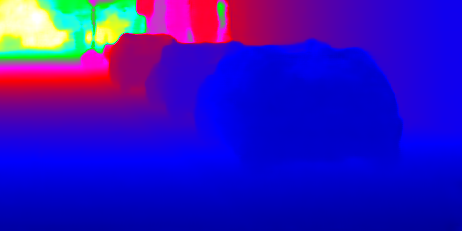} &
\includegraphics[width=0.45\columnwidth]{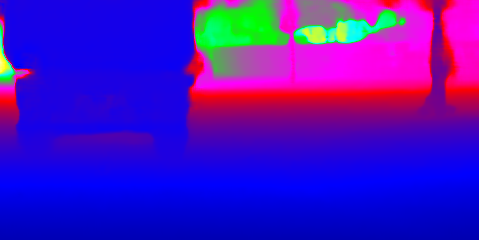} \\
& \includegraphics[width=0.45\columnwidth]{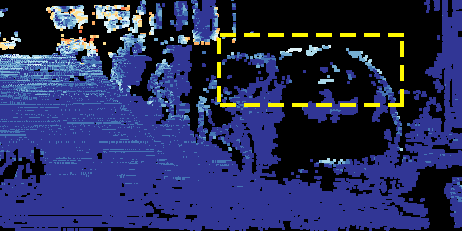} &
 \includegraphics[width=0.45\columnwidth]{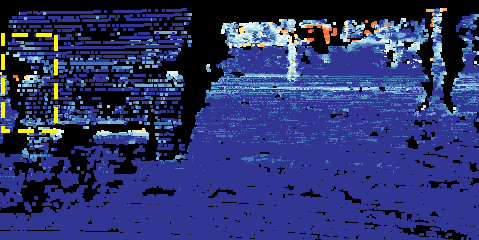} \\ 
\end{tabular}
\caption{Some qualitative examples for the top three performing methods from the KITTI-Depth dataset \cite{Uhrig2017} on the task of scene depth completion. \textbf{(a)} RGB input, \textbf{(b)} Our method \emph{MS-Net[LF]-L2 (gd)}, \textbf{(c)} Sparse-to-Dense (gd) \cite{mit} and \textbf{(d)} HMS-Net (gd) \cite{hms}. For each method, the top image is the prediction and the lower image is the error. Our method \emph{MS-Net[LF]-L2  (gd)} performs slightly better in handling outliers as highlighted with the yellow boxes, while Sparse-to-Dense produces smoother edges due to the use of a smoothness loss. Note that this figure is best viewed on screens.
}
\label{fig:qual}
\end{figure}

Figure \ref{fig:qual} shows some qualitative results for the top performing methods from the benchmark server. For our method, we show examples from \textcolor{black}{\emph{MS-Ne[LF]-L2 (gd)}} that was trained using the L2-norm loss, which achieved the lowest RMSE error. Generally, our method and the other two methods in comparison perform equally well with some minor differences. Our method performs better with tiny details (highlighted using the yellow boxes) such as the car edges in the first row and the poles far away in the second row. Our method was also able to remove outliers and highly sparse regions as shown on the third row. On the other hand, \emph{Sparse-to-Dense (gd)} produces smoother edges than our method and \emph{HMS-Net (gd)} due to the use of a smoothness loss that penalizes the second derivative of the prediction during training.

\renewcommand{\arraystretch}{0.5}

\color{black}
\subsection{The NYU-Depth-v2 Dataset Comparison}
On the NYU-Depth-v2 dataset, we compare \emph{MS-Net[LF]} and \emph{EncDec-Net[EF]} against published state-of-the-art methods that utilize the RGB image and \textcolor{black}{uniformly sampled pixels from the depth map as an input}. \emph{Sparse-to-dense} \cite{Ma2017SparseToDense} utilizes a  single deep regression network to learn a dense depth map from the RGB-D raw data input. \emph{Liao }\etal \cite{liao2017parse}  solve this problem by constructing a residual network which combines classification and regression losses to learn a dense depth map. \emph{Cheng} \etal \cite{cheng2018depth} proposed a convolutional spatial propagation network (CSPN), which learns the affinity matrix needed to predict a dense depth map.

Table \ref{tab:2} shows the quantitative results for the methods in comparison on the NYU-Depth-v2 dataset. 
%For very sparse depth input (50 samples), our encoder-decoder architecture \emph{EncDec-Net[EF]} achieves the best results despite having a much smaller number of parameters ($\approx484$k) compared to Sparse-to-Dense \cite{Ma2017SparseToDense} which has ($\approx31$M parameters). On the other hand, our multi-stream architecture \emph{MS-Net[LF]} achieves only average results, presumably due to the lack of multiple scales in \emph{MS-Net[LF]}, which are very useful in case of high degrees of sparsity. 
For a very sparse input (200 samples), our \emph{EncDec-Net[EF]} achieves the best results with a huge margin to other methods in comparison. In addition, \emph{MS-Net[LF]} achieving the second best results. On the other hand, Sparse-to-Dense \cite{Ma2017SparseToDense} performs significantly worse than our proposed method despite have two orders of magnitude larger number of parameters. For a denser input (500 samples), \emph{EncDec-Net[EF]} achieves the second best results with a small margin to \emph{UNet+CSPN} \cite{cheng2018depth}. \textcolor{black}{However, \cite{cheng2018depth} requires "Preserving Depth" values from the input in order to update the learned affinity between layers. This requirement is not always appropriate, \eg in case of corrupted/incorrect input in the KITTI-Depth dataset \cite{qiu2018deeplidar} due to occlusion.}

\renewcommand{\arraystretch}{1.3}
\begin{table}[]
\begin{center}
\begin{tabular}{L{3.0cm} | C{1.2cm} | C{0.8cm} C{0.8cm} C{0.8cm} C{0.8cm} C{0.8cm} }
\specialrule{.2em}{.1em}{.1em} 
 & \#Samples & RMSE & REL  & $\delta_1$ & $\delta_2$ & $\delta_3$ \\
\hline
%MS-Net[LF] \textbf{(ours)} & 50 & 0.354 & 0.074 &  92.5 & 97.9 & 99.3 \\
%Sparse-to-Dense \cite{Ma2017SparseToDense} & 50 & 0.281 & 0.059 &  95.5 & \textbf{99.0} & 99.6 \\
%EncDec-Net[EF] \textbf{(ours)} & 50 & \textbf{0.285} & \textbf{0.052}&  95.5 & \textit{98.8} & 99.6 \\
%\hline
Liao \etal \cite{liao2017parse} & 225 & 0.442& 0.104 &  87.8 & 96.4 & 98.9 \\
\hline
Sparse-to-Dense \cite{Ma2017SparseToDense} & 200 & 0.230 & 0.044 &  97.1 & 99.4 & 99.8 \\
MS-Net[LF] \textbf{(ours)} & \textcolor{black}{(0.28\%)} & \textit{0.192} & \textit{0.030} &  \textit{97.9} & \textit{99.5} &\textit{ 99.8} \\
EncDec-Net[EF] \textbf{(ours)} &  & \textbf{0.171} & \textbf{0.026} &  \textbf{98.3} & \textbf{99.6} & \textbf{99.9} \\
\hline
Sparse-to-Dense \cite{Ma2017SparseToDense} &  & 0.224 & 0.043 &  97.8 & 99.5 & 99.9 \\
SPN \cite{cheng2018depth} &  & {0.162} & 0.027 &  {98.5} & {99.7} & {99.9} \\
UNet+SPN \cite{cheng2018depth} & 500 & {0.144} & {0.022} &  {98.8} & {99.8} & 100.0 \\
CSPN \cite{cheng2018depth} & \textcolor{black}{(0.72\%)} & {0.136} & {0.021} &  {99.0} & {99.8} & 100.0 \\
MS-Net[LF] \textbf{(ours)} &  & 0.129 & 0.018 &  99.1 & 99.8 & 100.0 \\
EncDec-Net[EF] \textbf{(ours)} &  & \textit{0.123} & \textit{0.017} &  \textit{99.1} & \textit{99.8} & {100.0} \\
UNet+CSPN \cite{cheng2018depth} &  & \textbf{0.117} & \textbf{0.016} &  \textbf{99.2} & \textbf{99.9} & {100.0} \\
\specialrule{.2em}{.1em}{.1em} 
\end{tabular}
\end{center}
\caption{Quantitative results for methods in comparison on the NYU-Depth-v2 dataset \cite{nyu}. \textcolor{black}{\emph{\#Samples} states the number of depth pixels that were uniformly sampled and the sparsity levels are indicated in brackets.}} 
\label{tab:2}
\end{table}

\begin{figure}
\centering
\includegraphics[width=\columnwidth]{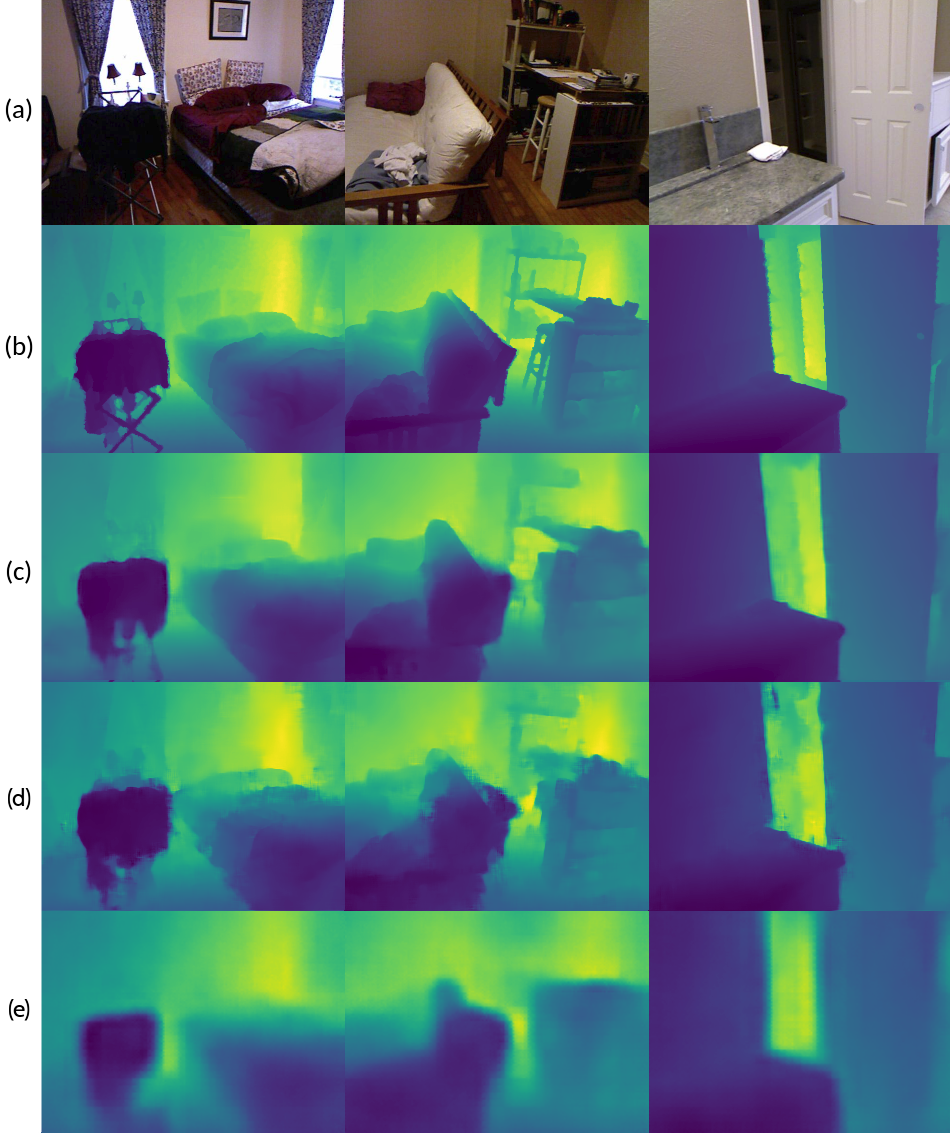}%
\caption{Some qualitative results on the NYU-Depth-v2 dataset with 200 randomly sampled depth samples as the input. \textbf{(a)} RGB input, \textbf{(b)} The groundtruth depth, \textbf{(c)} our \emph{EncDec-Net[EF]} results, \textbf{(d)} our \emph{MS-Net[LF]} results, and \textbf{(e)} Sparse-to-Dense \cite{Ma2017SparseToDense} results. }
\label{fig:nyu_qual}
\end{figure}

Figure \ref{fig:nyu_qual} shows some qualitative examples on the NYU-Depth-v2 dataset. Both our methods \emph{EncDec-Net[EF]} and \emph{MS-Net[LF]} produce remarkably better reconstructions of the dense map that Sparse-to-Dense \cite{Ma2017SparseToDense}, in particular with respect to edges sharpness and the level of details. The predictions from \cite{Ma2017SparseToDense} are very blurry and give a global depth estimation for local regions. However, \emph{EncDec-Net[EF]} yields smoother and more consistent reconstruction than \emph{MS-Net[LF]}, especially along edges.

\color{black}

\section{Analysis}
\label{sec:analys}
\color{black}

In this section, we analyze different components of our proposed method thoroughly. Since the NYU-Depth-v2 allows changing the degree of sparsity, we use it to evaluate our method's performance with varying degrees of sparsity. Other analyses are performed on the KITTI-Depth dataset.

\subsection{Varying Degree of Sparsity}
\begin{figure}[t]
\centering
\includegraphics[width=0.8\columnwidth]{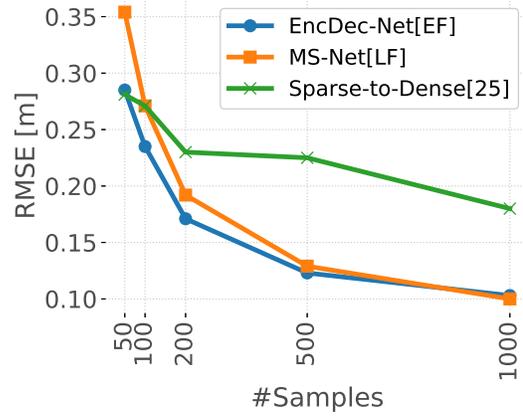}%
\caption{The effect of varying the degree of sparsity in the NYU-Depth-v2 dataset \cite{nyu} on our proposed method. \emph{EncDec-Net[EF]} performs very well with different degrees of sparsity, while \emph{EncDec-Net[EF]} performs slightly worse with high degrees of sparsity.}
\label{fig:sparsity}
\end{figure}

The NYU-Depth-v2 dataset allows changing the degree of sparsity by altering the number of depth samples provided at the input. Figure \ref{fig:sparsity} shows how our architectures \emph{MS-Net[LF]} and \emph{EncDec-Net[EF]} perform with varying degrees of sparsity. \emph{EncDec-Net[EF]} performs very well with different degrees of sparsity even with a very sparse input ($\sim 0.01\%$). This is due to the use of multiple scales, which allows exploiting depth information at different scales. On the other hand, \emph{MS-Net[LF]} performs worse as it has only one scale level. However, when the sparsity degree decreases, i.e. the number of depth samples is increased, \emph{MS-Net[LF]} approaches \emph{EncDec-Net[EF]} until they produce very similar results at lower degrees of sparsity. Sparse-to-dense \cite{Ma2017SparseToDense} performs very well at very sparse input. However, with the decreasing level of sparsity, it does not seem that the network is significantly benefiting from the additional depth samples, contrarily to our method.

\color{black}

\subsection{The Choice of the Non-negative Function}
\begin{figure}[t]
\centering
\subfloat[]{\includegraphics[width=0.49\columnwidth]{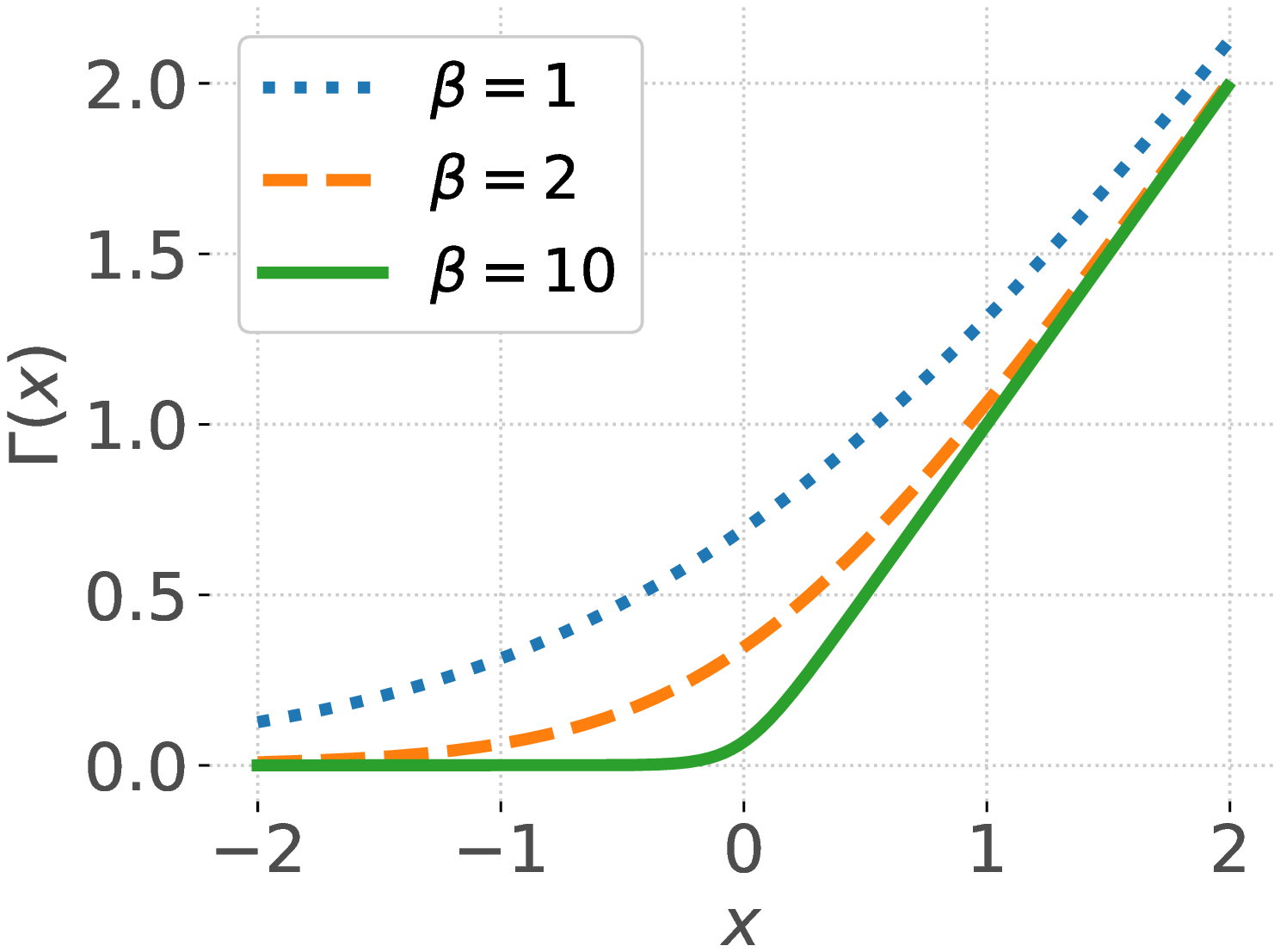}%
\label{fig:softplus_beta}}
\hfil
\subfloat[]{\includegraphics[width=0.49\columnwidth]{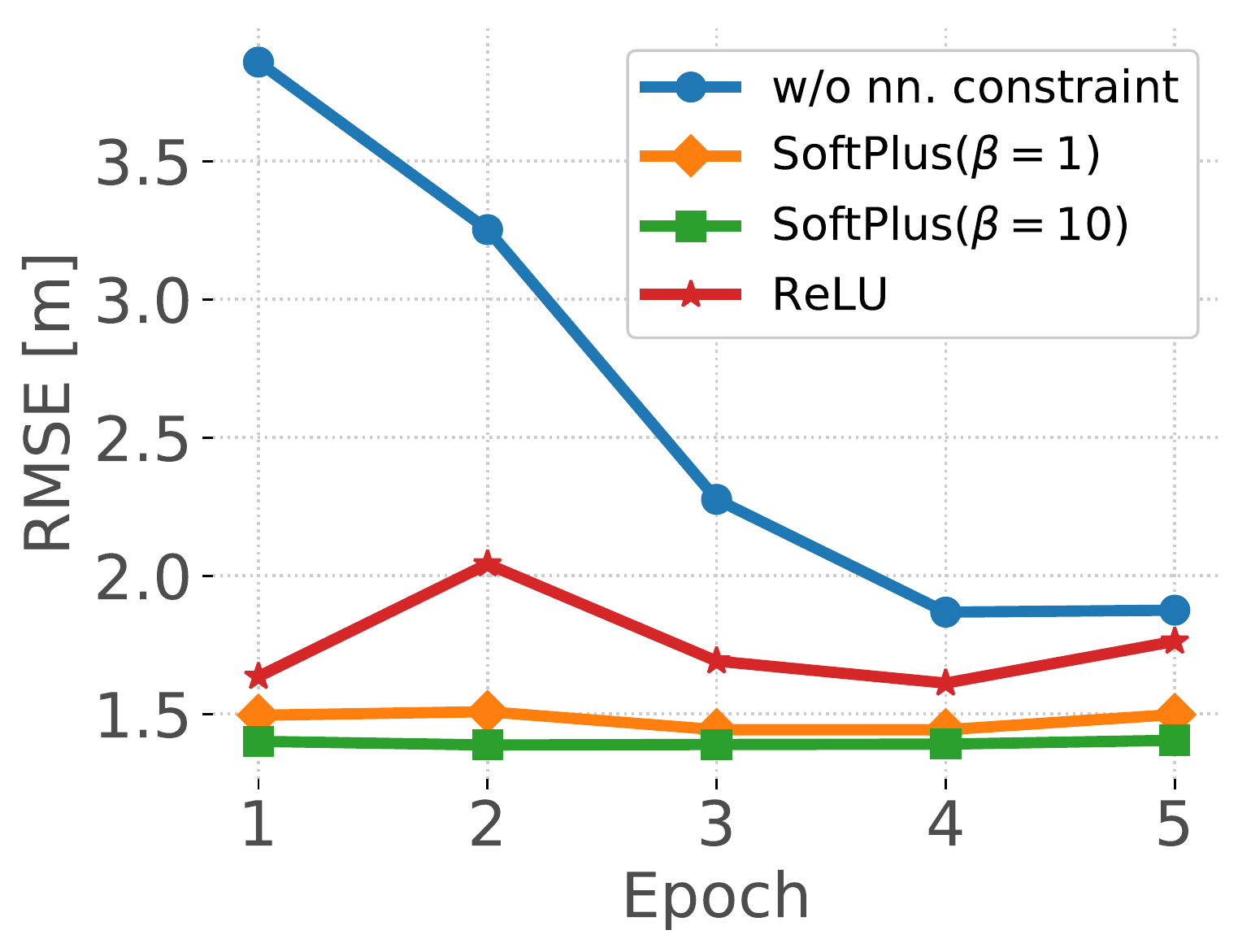}%
\label{fig:nn_const_conv}}
\caption{\textbf{(a)} The SoftPlus function with different scaling factors. At $\beta=10$, the SoftPlus function gives a good differentiable approximation of the ReLU function. \textbf{(b)} Convergence curves for our unguided network with the presence and absence of the non-negativity constraint.}
\label{fig:nn}
\end{figure}

The choice of the non-negative function mainly depends on the desired characteristics. 
The most obvious choice is to clamp non-negative values as in the ReLU function. However, the ReLU has problems with the discontinuous derivative. Therefore, we consider a good continuous approximation for the ReLU, the SoftPlus function $\Gamma(x) = \frac{1}{\beta} \ \log(1+\exp(\beta  x)$. With the right choice of the $\beta$, we can have a very good approximation of the ReLU function, \textit{e.g.} when $\beta=10$, as shown in Figure \ref{fig:softplus_beta}. The derivative of the SoftPlus function is continuous which gives more flexibility to the network during training. Figure \ref{fig:nn_const_conv} shows how the choice of the non-negative function affects the convergence of the network. The ReLU function shows poor convergence and keeps fluctuating as the derivatives are not continuous and the network struggles to converge. SoftPlus on the other hand converges very fast due to the continuous derivative and with the right choice of $\beta$, the results are improved.

\subsection{The Non-negativity Constraint Impact}

\renewcommand{\arraystretch}{1.3}
\begin{table}[b]
\begin{center}
\begin{tabular}{L{3.5cm} | C{1.2cm} C{1.2cm} C{1cm} C{1cm} }
\specialrule{.2em}{.1em}{.1em} 
 Method \newline [Non-Negativity Function]& MAE [mm] & RMSE [mm] & iMAE [1/km] & iRMSE [1/km] \\
\hline
\textcolor{black}{MS-Net[LF]-L1} (gd) [SoftPlus($\beta=10$)] &  \textbf{209.56} & \textbf{908.76} & \textbf{0.90} &\textbf{2.50} \\ %
\textcolor{black}{MS-Net[LF]-L1} (gd) [N/A] &  277.67 & 1042.65 & 1.28 & 3.50 \\
\specialrule{.2em}{.1em}{.1em} 
\end{tabular}
\end{center}
\caption{The impact of enforcing non-negativity on normalized convolution layers. \emph{NConv-CNN-L1 (gd)} with SoftPlus achieves significantly better results than the case without enforcing non-negativity. The results shown are for the  \emph{selected validation} set of the KITTI-Depth dataset \cite{Uhrig2017}.}
\label{tab:nn_const}
\end{table}

To study the effect of enforcing non-negativity constraints on our trained filters, we compare the convergence of our proposed unguided normalized convolution module described in section \ref{sec:ncnn} in the presence and the absence of the non-negativity enforcement. Since our proposed confidence measure cannot be used in the absence of the non-negativity constraints, we propagate confidences by applying a max pooling operations on the confidence map as in \cite{Uhrig2017, hms, ncnn}. Besides, only the data error term in our proposed loss in (\ref{eq:13}) is used because of the absence of output confidence. Both networks were trained on 10k training images for 5 epochs with a constant learning rate of 0.01. The networks were trained on the disparity instead of depth since both networks perform better when trained on disparity. Figure \ref{fig:nn_const_conv} shows the convergence curves for both networks. When enforcing the non-negativity constraints, the network converges after 1 epoch, while the other network not enforcing the non-negativity constraint starts to converge after 4 epochs and to a higher error value. This demonstrates that our proposed non-negativity constraint helps the network to converge faster and to achieve significantly better results.

The overall effect of the non-negativity constraints on the guided network is shown in Table \ref{tab:nn_const}. Discarding the non-negativity constraint significantly degrades the results with respect to all evaluation metrics. This is potentially caused by the lack of guidance provided by the output confidence or by the the poor estimation of depth produced by the unguided network.

\subsection{The Impact of the Proposed Loss}
\begin{figure}[t]
\centering
\includegraphics[width=0.9\columnwidth]{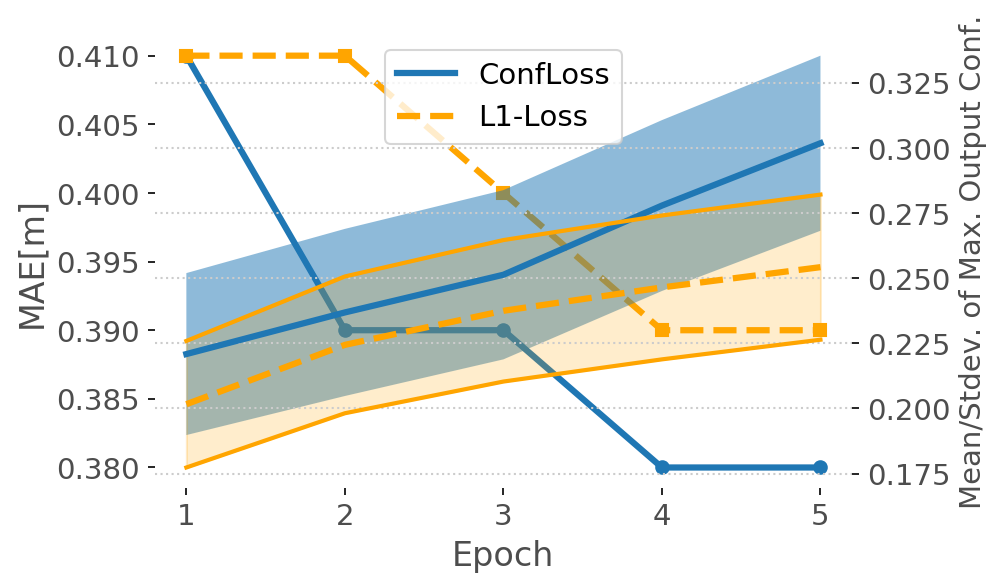}%
\caption{The impact of the proposed loss on confidence levels. The right axis represents the mean and standard deviation of maximum output confidence value over all images, while the left axis has the MAE in meters. When using a loss with only a data term (Huber norm Loss), output confidence levels are lower, while our proposed loss achieves monotonically increasing confidence levels as well as a lower MAE. Note that the shaded area represents the standard deviation. \textcolor{black}{The results shown are for the \emph{selected validation} set of the KITTI-Depth dataset \cite{Uhrig2017}.}}
\label{fig:confloss}
\end{figure}

To study the impact of the confidence term in our proposed loss in (\ref{eq:13}), we train our unguided normalized convolution network described in section \ref{sec:ncnn} twice: once using the proposed loss function with confidence term and once using only the Huber norm loss (\ref{eq:12}). Figure \ref{fig:confloss} shows the mean and the standard deviation of the maximum output confidence over images in the selected validation set on the right axis and the MAE error on the left axis. The network trained with our proposed loss produces a monotonically increasing confidence map while improving the data error until convergence. On the other hand, the network trained with only the Huber norm loss has lower levels of output confidence in general and it also converges to a higher MAE.

\subsection{The Learned Filters}

\begin{figure}[t]
\centering
\includegraphics[width=0.8\columnwidth]{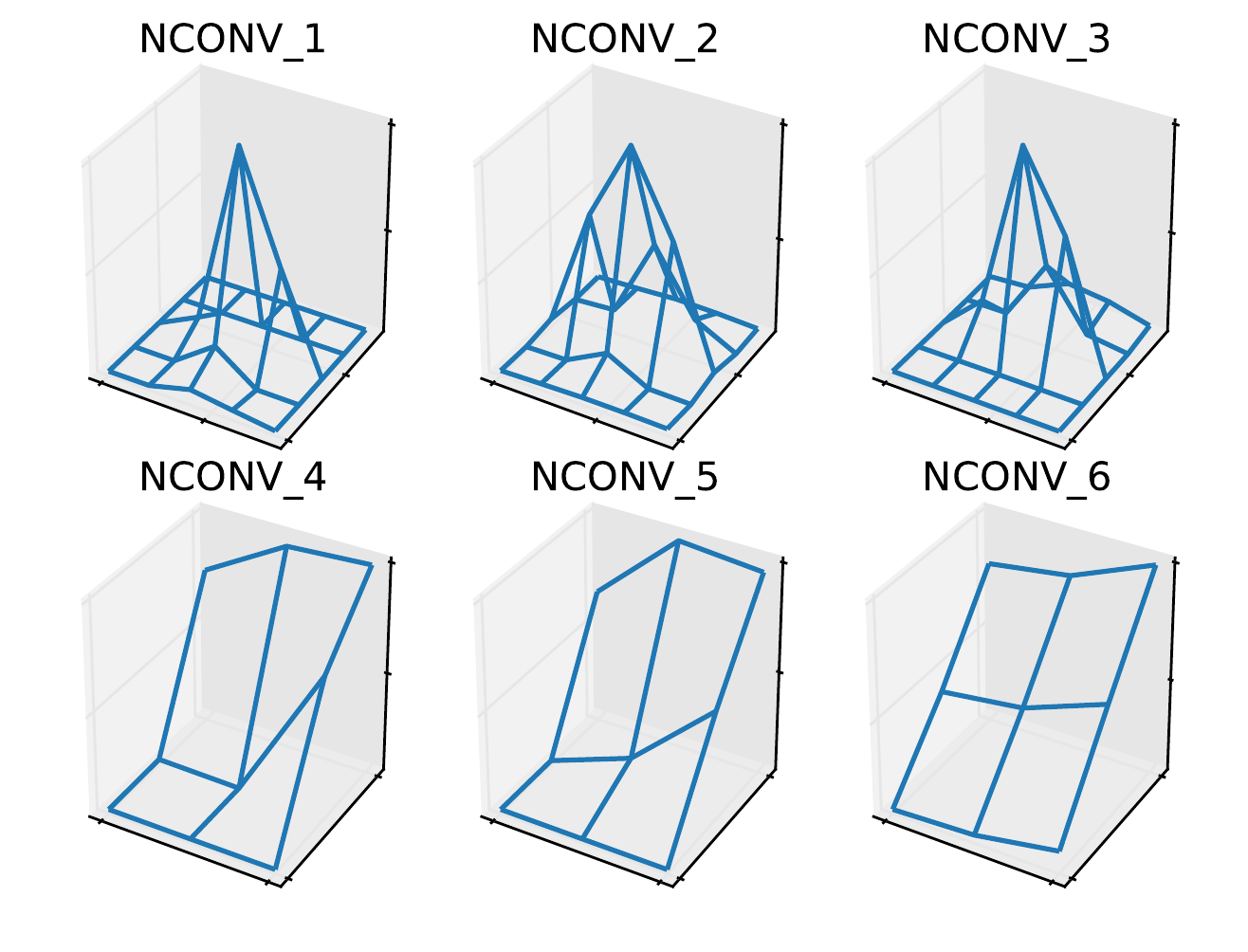}%
\caption{A visualization of the learned filters from our proposed unguided normalized convolution network \textcolor{black}{on the KITTI-Depth dataset \cite{Uhrig2017}.} Layer names match their correspondences in Figure \ref{fig:ncnn}.}
\label{fig:trained_filters}
\end{figure}

The unguided normalized convolution network acts as a multi-scale generic estimator for the data. During training, this generic estimator is learned from the data using back-propagation. Some examples of the learned filters are shown in Figure \ref{fig:trained_filters}. The first row of the figure shows some of the learned filters for layers NCONV[1-3], which are asymmetric low-pass filters. Those filters attempt to construct the missing pixels from their neighborhood. On the other hand, the second row of the figure shows the learned filters for layers NCONV[4-6], which resemble linear ramps. Those filters try to scale the output from each scale for an efficient fusion with other scales. %This finding means that the weights of the normalized convolution layer could be initialized with a 2D Gaussian kernel with additive noise. This would lead to a faster convergence for the network. 

\subsection{Guided Normalized CNN Ablation Study}
\renewcommand{\arraystretch}{1.3}
\begin{table}[b]
\begin{center}
\begin{tabular}{L{3.0cm}| C{0.5cm} | C{1.2cm} C{1.2cm} C{1cm} C{1cm} }
\specialrule{.2em}{.1em}{.1em} 
& DS & MAE [mm] & RMSE [mm] & iMAE [1/km] & iRMSE [1/km] \\
\hline
Baseline & K & \textbf{233.25} & \textbf{870.82} & \textbf{1.03} & \textbf{2.75} \\ % Exp_3ffull4
Baseline (end-to-end) & & 244.98 & 886.09 & 1.11 & 3.02 \\ % Exp_3ffull5
Baseline w/o DR & &  245.11 & 912.82 & 1.10 & 3.03\\ % Exp_3ffull7
Baseline w/o OC & & 244.87 & 919.55 & 1.09 & 2.97 \\ % Exp_3ffull9
\hline
\textcolor{black}{EncDec-Net[EF]} & N & 236.83 & 1007.71 & 0.99 & 2.75 \\ % Exp_4c
\textcolor{black}{EncDec-Net[EF]$\times2$ w/o NConv} & & 274.73 & 1122.51 & 1.19 & 3.28 \\
\specialrule{.2em}{.1em}{.1em} 
\end{tabular}
\end{center}
\caption{\textcolor{black}{\textit{DS} refers to the used dataset, \textit{K} is the KITTI-Depth, \textit{N} is the NYU-Depth-v2. On the KITTI-Depth dataset, baseline refers to \textcolor{black}{\emph{MS-Net[LF]-L2 (gd)}}, \emph{DR} refers to the Depth Refinement layers as indicated in Figure \ref{fig:intro} and \emph{OC} is the use of the output confidence in the RGB feature extraction network. When the baseline is trained end-to-end, the performance is slightly degraded. The depth refinement layers also contribute to the results. Discarding the output confidence, degrades the results. On the NYU-Depth-v2, a network with standard convolution and double number of layers fails to achieve comparable results to EncDec-Net[EF]. }}
\label{tab:abl}
\end{table}

In this section, we study the effect of different components of our best performing guided network. Since the benchmarks is ranked based on the RMSE, we use \textcolor{black}{\emph{MS-Net[LF]-L2 (gd)}} as a baseline. When the whole network was trained end-to-end, the results are slightly degraded as shown in Table \ref{tab:abl}. This could be a result of vanishing gradients since the network becomes deeper. Removing either the depth refinement layers or the output confidence has almost the same influence on the performance of the network as shown in Table \ref{tab:abl}. The reason might be that the depth refinement layers contribute to handling some outliers that violate the estimation from the unguided network. On the other hand, the use of the output confidence provides the RGB stream with information about regions with low confidence that are highly likely to contribute to the error. Therefore, discarding the output confidence increases the error by approximately $10\%$, which demonstrates its contribution.

\textcolor{black}{To further validate our proposed architecture, we perform an experiment on the KITTI-Depth dataset by increasing the number of network layers of \emph{EncDec-Net[EF]} by a factor of 2 and removing both our confidence propagation and normalization components. We denoted this experiment as \emph{EncDec-Net[EF]$\times2$ w/o NConv} and Table \ref{tab:abl} shows  that the resulting large network provides an 1122.51 [mm] RMSE score which is inferior to 1007.71 [mm] achieved by our proposed light-weight architecture with confidence propagation and normalization.}

%\textcolor{red}{To demonstrate the overall impact of the normalized convolution layers and confidence propagation, we evaluate a variation of \emph{EncDec-Net[EF]} network by removing the unguided normalized convolution network and accordingly the output confidence while doubling the number of layers to be 20 instead of 10. We denote it as \emph{EncDec-Net[EF]$\times2$ w/o NConv} and Table \ref{tab:abl} shows that it still achieves significantly worse results than our proposed \emph{EncDec-Net[EF]}.}

\subsection{Number of Parameters and Runtime Comparison}
In this section, we compare the number of parameters and the runtime for some of the methods in comparison. For the KITTI-Depth dataset, the number of parameters is calculated from the network descriptions in the related papers, while the runtime is taken from the benchmark server \cite{Uhrig2017}. Table \ref{tab:param} shows that our unguided network \emph{NConv-CNN (d)} has the lowest number of parameters and runtime compared to all other unguided methods, which makes it most suitable for embedded applications with limited computational resources. We maintain the low number of parameters in our guided network \textcolor{black}{\emph{MS-Net-L2[LF] (gd)}} that has 356k parameters, which is at least one order of magnitude fewer than all other guided methods in the comparison. This \textcolor{black}{leads} to the lowest runtime of 0.02 seconds among the guided methods which satisfies real-time constraints and has a high potential to maintain the real-time performance if evaluated on embedded devices. The huge decrease in the number of parameters did not degrade the results as was shown in the quantitative results earlier, which demonstrates the efficiency of our proposed method. Note that the runtime between our unguided and guided network do not scale linearly as our unguided network includes many time consuming operations such as downsampling, slicing and upsamping, while the guided network adds only convolution operations.

\textcolor{black}{For the NYU-Depth-v2 dataset, our method has a drastically lower number of parameters compared to Sparse-to-Dense \cite{Ma2017SparseToDense} ($\sim 1\%$ the number of parameters). However, our method still managed to achieve better results at different levels of sparsity.}

\renewcommand{\arraystretch}{1.3}
\begin{table}[t]
\begin{center}
\begin{tabular}{L{4.0cm} | C{2.0cm} C{2.0cm}}
\specialrule{.2em}{.1em}{.1em} 
 & \#Params & Runtime [sec] \\
\hline
Sparse-to-Dense (d) \cite{mit} & $5.53 \times 10^6$ & 0.04	\\
SparseConv \cite{Uhrig2017} & $2.5 \times 10^4 $ & \textbf{0.01} \\
ADNN \cite{Chodosh2018} & $1.7 \times 10^3$ & 0.04	\\
NConv-CNN (d) \textbf{(ours)} & \textcolor{black}{$\mathbf{4.8 \times 10^2}$} & $\mathbf{0.01}$ \\
\hline
Sparse-to-Dense (gd) \cite{mit} & $5.54 \times 10^6$ & 0.08	\\
Spade (gd) \cite{valeo} & $ \sim 5.3 \times 10^6 $ & 0.07 \\
\textcolor{black}{MS-Net[LF]-L2} (gd) \textbf{(ours)} & $\mathbf{3.56 \times 10^5}$ & $\mathbf{0.02}$ \\
\hline
Sparse-to-Dense \cite{Ma2017SparseToDense} & $3.18 \times 10^7$ & 0.01 \\
\textcolor{black}{EncDec-Net[EF]} (gd) \textbf{(ours)} & $\mathbf{4.84 \times 10^5}$ & 0.01 \\
\specialrule{.2em}{.1em}{.1em} 
\end{tabular}
\end{center}
\caption{Number of parameters and runtime for some methods in comparison (lower is better). The upper section is for unguided networks, the middle section is for guided networks and the lower section is for the NYU-Depth-v2 experiments. Note that the exact number of \emph{Spade (gd) \cite{valeo}} is not mentioned in the paper, so we give the number of parameters for the NASNet \cite{nasnet} that they utilize.}
%\caption{Number of parameters and runtime for some methods in comparison (lower is better). The upper section is for unguided networks. Our unguided network \emph{NConv-CNN (d)} has a significantly lower number of parameters than all other unguided networks. The middle section is for guided networks. Our guided network \textcolor{red}{\emph{MS-Net[LF]-L2 (gd)}} has at least one order of magnitude fewer parameters than all other guided methods. Both our unguided and guided networks have the lowest runtime due to the small number of parameters. Note that the exact number of \emph{Spade (gd) \cite{valeo}} is not mentioned in the paper, so we give the number of parameters for the NASNet \cite{nasnet} that they utilize. The lower section is for the NYU-Depth-v2 experiments. \textcolor{red}{Our encoder decoder network \emph{EncDec-Net[EF]} has the lowest number of parameters with at least two orders of magnitude than other methods.}}
\label{tab:param}
\end{table}

\color{black}
\subsection{Output Confidence/Error Correlation }
We have shown empirically that the output confidence is useful to improve the results.
To gain further understanding, we perform a correlation analysis between the prediction absolute error and the negative 
logarithm of the output confidence 
(similar to the log likelihood). 
\newline

\noindent We employ Pearson's correlation measure defined as:
\begin{equation}
\rho_{X,Y} = \frac{\text{cov}(X,Y)}{\sigma_X \sigma_Y} \enspace .
\end{equation}

The analysis is performed on the output from our unguided network \emph{NConv-CNN (d)} both on the KITTI-Depth and the NYU-Depth-v2 datasets. Since the distributions for the error and the output confidence are unknown, we perform histogram equalization and transform the error values and confidences accordingly. 

\begin{figure}[t]
	\centering
	\subfloat[]{\includegraphics[width=0.49\columnwidth]{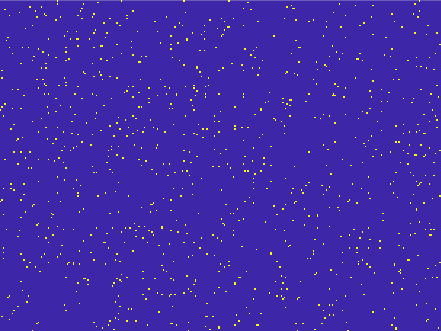}%
		\label{fig:conf_baseline_inp}}
	\hfil
	\subfloat[]{\includegraphics[width=0.49\columnwidth]{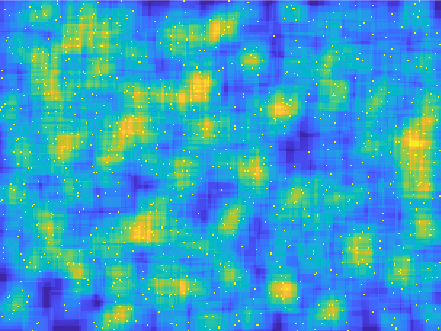}%
		\label{fig:conf_baseline_out}}
	\caption{An illustration of the baseline for output confidence/error correlation on the NYU-Depth-v2 dataset. \textbf{(a)} The binary input confidence map. \textbf{(b)} Interpolated input confidence map using the normalized convolution with the naive basis and a Gaussian applicability. Confidences are maximal at input points location and are increasingly attenuated the further we move from the input points.}
	\label{fig:conf_baseline}
\end{figure}

To form a baseline, we produce a \emph{naive} output confidence by interpolating the input confidence mask using the normalized convolution with the naive basis and a Gaussian applicability. This produces high confidence at location where the input is valid, and increasingly attenuated confidences the further we move from input point locations as illustrated in Figure \ref{fig:conf_baseline_out}.
The baseline gives an average Pearson's correlation measure of 0.1 on the NYU-Depth-v2 and \mbox{-~0.25} on the KITTI-Depth validation sets. We attribute this negative correlation to the faulty input in the KITTI-Depth dataset that does not match the groundtruth \cite{qiu2018deeplidar}.

Contrarily, our proposed output confidence achieves a significantly higher correlation of 0.3 and 0.4. This correlation is considerably high as the correlation upperbound for unknown distributions is low.

\color{black}

\section{Conclusion}
\label{sec:concl}
In this paper, we have proposed a normalized convolution layer for unguided scene depth completion on highly sparse data by treating the validity masks as a continuous confidence field. We proposed a method to propagate confidences between CNN layers. This enabled us to produce a point-wise continuous confidence map for the output from the deep network. For fast convergence, we algebraically constrained the learned filters to be non-negative, acting as a weighting function for the neighborhood. Furthermore, a loss function is proposed that simultaneously minimizes the data error and maximizes the output confidence. Finally, we proposed a fusion strategy to combine depth and RGB information in our normalized convolution network framework for incorporating structural information. We performed comprehensive experiments on the KITTI-Depth benchmark, the NYU-Depth-v2 dataset and achieved superior performance with significantly fewer network parameters compared to the state-of-the-art.

\ifCLASSOPTIONcompsoc
  \section*{Acknowledgments}
\else
  \section*{Acknowledgment}
\fi

This research is funded by Vinnova through grant CYCLA, the Swedish Research Council through project grant 2018-04673, and VR starting grant (2016-05543).

\ifCLASSOPTIONcaptionsoff
  \newpage
\fi



\bibliographystyle{IEEEtran}
\bibliography{IEEEabrv,refs}

\begin{thebibliography}{10}
\providecommand{\url}[1]{#1}
\csname url@samestyle\endcsname
\providecommand{\newblock}{\relax}
\providecommand{\bibinfo}[2]{#2}
\providecommand{\BIBentrySTDinterwordspacing}{\spaceskip=0pt\relax}
\providecommand{\BIBentryALTinterwordstretchfactor}{4}
\providecommand{\BIBentryALTinterwordspacing}{\spaceskip=\fontdimen2\font plus
\BIBentryALTinterwordstretchfactor\fontdimen3\font minus
  \fontdimen4\font\relax}
\providecommand{\BIBforeignlanguage}[2]{{%
\expandafter\ifx\csname l@#1\endcsname\relax
\typeout{** WARNING: IEEEtran.bst: No hyphenation pattern has been}%
\typeout{** loaded for the language `#1'. Using the pattern for}%
\typeout{** the default language instead.}%
\else
\language=\csname l@#1\endcsname
\fi
#2}}
\providecommand{\BIBdecl}{\relax}
\BIBdecl

\bibitem{Uhrig2017}
\BIBentryALTinterwordspacing
J.~Uhrig, N.~Schneider, L.~Schneider, U.~Franke, T.~Brox, and A.~Geiger,
  ``{Sparsity Invariant CNNs},'' aug 2017. [Online]. Available:
  \url{http://arxiv.org/abs/1708.06500}
\BIBentrySTDinterwordspacing

\bibitem{ren2015shepard}
J.~S. Ren, L.~Xu, Q.~Yan, and W.~Sun, ``Shepard convolutional neural
  networks,'' in \emph{Advances in Neural Information Processing Systems},
  2015, pp. 901--909.

\bibitem{Chodosh2018}
\BIBentryALTinterwordspacing
N.~Chodosh, C.~Wang, and S.~Lucey, ``{Deep Convolutional Compressed Sensing for
  LiDAR Depth Completion},'' mar 2018. [Online]. Available:
  \url{http://arxiv.org/abs/1803.08949}
\BIBentrySTDinterwordspacing

\bibitem{Liu2018}
\BIBentryALTinterwordspacing
G.~Liu, F.~A. Reda, K.~J. Shih, T.-C. Wang, A.~Tao, and B.~Catanzaro, ``{Image
  Inpainting for Irregular Holes Using Partial Convolutions},'' apr 2018.
  [Online]. Available: \url{http://arxiv.org/abs/1804.07723}
\BIBentrySTDinterwordspacing

\bibitem{ncnn}
J.~Hua and X.~Gong, ``A normalized convolutional neural network for guided
  sparse depth upsampling.'' in \emph{IJCAI}, 2018, pp. 2283--2290.

\bibitem{hms}
Z.~{Huang}, J.~{Fan}, S.~{Yi}, X.~{Wang}, and H.~{Li}, ``{HMS-Net: Hierarchical
  Multi-scale Sparsity-invariant Network for Sparse Depth Completion},''
  \emph{ArXiv e-prints}, Aug. 2018.

\bibitem{valeo}
M.~Jaritz, R.~de~Charette, E.~Wirbel, X.~Perrotton, and F.~Nashashibi, ``Sparse
  and dense data with cnns: Depth completion and semantic segmentation,''
  \emph{arXiv preprint arXiv:1808.00769}, 2018.

\bibitem{mit}
F.~{Ma}, G.~{Venturelli Cavalheiro}, and S.~{Karaman}, ``{Self-supervised
  Sparse-to-Dense: Self-supervised Depth Completion from LiDAR and Monocular
  Camera},'' \emph{ArXiv e-prints}, Jul. 2018.

\bibitem{Ronneberger2015}
\BIBentryALTinterwordspacing
O.~Ronneberger, P.~Fischer, and T.~Brox, ``{U-Net: Convolutional Networks for
  Biomedical Image Segmentation}.''\hskip 1em plus 0.5em minus 0.4em\relax
  Springer, Cham, oct 2015, pp. 234--241. [Online]. Available:
  \url{http://link.springer.com/10.1007/978-3-319-24574-4{\_}28}
\BIBentrySTDinterwordspacing

\bibitem{nyu}
P.~K. Nathan~Silberman, Derek~Hoiem and R.~Fergus, ``Indoor segmentation and
  support inference from rgbd images,'' in \emph{ECCV}, 2012.

\bibitem{bmvc}
A.~Eldesokey, M.~Felsberg, and F.~S. Khan, ``Propagating confidences through
  cnns for sparse data regression,'' in \emph{The British Machine Vision
  Conference (BMVC), Northumbria University, Newcastle upon Tyne, England, UK,
  3-6 September, 2018}, 2018.

\bibitem{6335696}
N.~Yang, Y.~Kim, and R.~Park, ``Depth hole filling using the depth distribution
  of neighboring regions of depth holes in the kinect sensor,'' in \emph{2012
  IEEE International Conference on Signal Processing, Communication and
  Computing (ICSPCC 2012)}, Aug 2012, pp. 658--661.

\bibitem{7003768}
Y.~Shen, J.~Li, and C.~Lü, ``Depth map enhancement method based on joint
  bilateral filter,'' in \emph{2014 7th International Congress on Image and
  Signal Processing}, Oct 2014, pp. 153--158.

\bibitem{6865733}
Y.~Chiu, J.~Leou, and H.~Hsiao, ``Super-resolution reconstruction for kinect 3d
  data,'' in \emph{2014 IEEE International Symposium on Circuits and Systems
  (ISCAS)}, June 2014, pp. 2712--2715.

\bibitem{7995866}
S.~Wirges, B.~Roxin, E.~Rehder, T.~Kühner, and M.~Lauer, ``Guided depth
  upsampling for precise mapping of urban environments,'' in \emph{2017 IEEE
  Intelligent Vehicles Symposium (IV)}, June 2017, pp. 1140--1145.

\bibitem{konno2015intensity}
Y.~Konno, Y.~Monno, D.~Kiku, M.~Tanaka, and M.~Okutomi, ``Intensity guided
  depth upsampling by residual interpolation,'' in \emph{The Abstracts of the
  international conference on advanced mechatronics: toward evolutionary fusion
  of IT and mechatronics: ICAM 2015.6}.\hskip 1em plus 0.5em minus 0.4em\relax
  The Japan Society of Mechanical Engineers, 2015, pp. 1--2.

\bibitem{Knutsson}
H.~Knutsson and C.-F. Westin, ``Normalized and differential convolution,'' in
  \emph{Computer Vision and Pattern Recognition, 1993. Proceedings CVPR'93.,
  1993 IEEE Computer Society Conference on}.\hskip 1em plus 0.5em minus
  0.4em\relax IEEE, 1993, pp. 515--523.

\bibitem{farneback:phd_thesis}
G.~Farneb{\"a}ck, ``Polynomial expansion for orientation and motion
  estimation,'' Ph.D. dissertation, Link{\"o}ping University Electronic Press,
  2002.

\bibitem{pham2003normalized}
T.~Q. Pham and L.~J. Van~Vliet, ``Normalized averaging using adaptive
  applicability functions with applications in image reconstruction from
  sparsely and randomly sampled data,'' in \emph{Scandinavian Conference on
  Image Analysis}.\hskip 1em plus 0.5em minus 0.4em\relax Springer, 2003, pp.
  485--492.

\bibitem{Westelius302463}
C.-J. Westelius, ``Focus of attention and gaze control for robot vision,''
  Ph.D. dissertation, Link\"oping University, Computer Vision, The Institute of
  Technology, 1995.

\bibitem{Karlholm302807}
J.~Karlholm, ``Local signal models for image sequence analysis,'' Ph.D.
  dissertation, Link\"oping University, Computer Vision, The Institute of
  Technology, 1998.

\bibitem{huber1964robust}
P.~J. Huber \emph{et~al.}, ``Robust estimation of a location parameter,''
  \emph{The annals of mathematical statistics}, vol.~35, no.~1, pp. 73--101,
  1964.

\bibitem{zeiler2014visualizing}
M.~D. Zeiler and R.~Fergus, ``Visualizing and understanding convolutional
  networks,'' in \emph{European conference on computer vision}.\hskip 1em plus
  0.5em minus 0.4em\relax Springer, 2014, pp. 818--833.

\bibitem{kitti}
A.~Geiger, P.~Lenz, C.~Stiller, and R.~Urtasun, ``Vision meets robotics: The
  kitti dataset,'' \emph{International Journal of Robotics Research (IJRR)},
  2013.

\bibitem{Ma2017SparseToDense}
F.~Ma and S.~Karaman, ``Sparse-to-dense: Depth prediction from sparse depth
  samples and a single image,'' \emph{arXiv preprint arXiv:1709.07492}, 2017.

\bibitem{cheng2018depth}
X.~Cheng, P.~Wang, and R.~Yang, ``Depth estimation via affinity learned with
  convolutional spatial propagation network,'' in \emph{Proceedings of the
  European Conference on Computer Vision (ECCV)}, 2018, pp. 103--119.

\bibitem{eigen2014depth}
D.~Eigen, C.~Puhrsch, and R.~Fergus, ``Depth map prediction from a single image
  using a multi-scale deep network,'' in \emph{Advances in neural information
  processing systems}, 2014, pp. 2366--2374.

\bibitem{ipbasic}
J.~Ku, A.~Harakeh, and S.~L. Waslander, ``In defense of classical image
  processing: Fast depth completion on the cpu,'' \emph{arXiv preprint
  arXiv:1802.00036}, 2018.

\bibitem{liao2017parse}
Y.~Liao, L.~Huang, Y.~Wang, S.~Kodagoda, Y.~Yu, and Y.~Liu, ``Parse geometry
  from a line: Monocular depth estimation with partial laser observation,'' in
  \emph{2017 IEEE International Conference on Robotics and Automation
  (ICRA)}.\hskip 1em plus 0.5em minus 0.4em\relax IEEE, 2017, pp. 5059--5066.

\bibitem{qiu2018deeplidar}
J.~Qiu, Z.~Cui, Y.~Zhang, X.~Zhang, S.~Liu, B.~Zeng, and M.~Pollefeys,
  ``Deeplidar: Deep surface normal guided depth prediction for outdoor scene
  from sparse lidar data and single color image,'' \emph{arXiv preprint
  arXiv:1812.00488}, 2018.

\bibitem{nasnet}
B.~Zoph, V.~Vasudevan, J.~Shlens, and Q.~V. Le, ``Learning transferable
  architectures for scalable image recognition.''

\end{thebibliography}

\newpage
\begin{IEEEbiography}[{\includegraphics[width=1in,height=1.25in,clip,keepaspectratio]{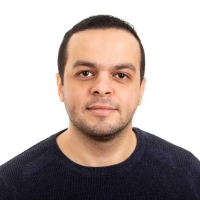}}]{Abdelrahman Eldesokey}
is a Ph.D. student at the Computer Vision Laboratory, Link\"oping University, Sweden. . He received his M.Sc. degree in Informatics from Nile University, Egypt in 2016.  He is also affiliated with the Wallenberg AI, Autonomous Systems and Software Program (WASP). His research interests include deep learning for computer vision and autonomous driving with a focus on uncertain and sparse data. %He is also interested in classical geometry for vision such as ellipse fitting. 
\end{IEEEbiography}

\begin{IEEEbiography}[{\includegraphics[width=1in,height=1.25in,clip,keepaspectratio]{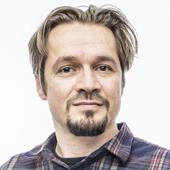}}]{Michael Felsberg}
received the Ph.D. degree in engineering from the University of Kiel, Kiel, Germany, in 2002. Since 2008, he has been a Full Professor and the Head of the Computer Vision Laboratory, Link\"oping University, Link\"oping, Sweden. His current research interests include signal processing methods for image analysis, computer and robot vision, and machine learning. He has published more than 100 reviewed conference papers, journal articles, and book contributions. He was a recipient of awards from the German Pattern Recognition Society in 2000, 2004, and 2005, from the Swedish Society for Automated Image Analysis in 2007 and 2010, from Conference on Information Fusion in 2011 (Honorable Mention), and from the CVPR Workshop on Mobile Vision 2014. He has achieved top ranks on various challenges (VOT: 3rd 2013, 1st 2014, 2nd 2015; VOT-TIR: 1st 2015; OpenCV Tracking: 1st 2015; KITTI Stereo Odometry: 1st 2015, March). He has coordinated the EU projects COSPAL and DIPLECS, he is an Associate Editor of the Journal of Mathematical Imaging and Vision, Journal of Image and Vision Computing, Journal of Real-Time Image Processing, Frontiers in Robotics and AI. He was Publication Chair of the International Conference on Pattern Recognition 2014 and Track Chair 2016, he was the General Co-Chair of the DAGM symposium in 2011, and he will be general Chair of CAIP 2017.
\end{IEEEbiography}

\begin{IEEEbiography}[{\includegraphics[width=1in,height=1.25in,clip,keepaspectratio]{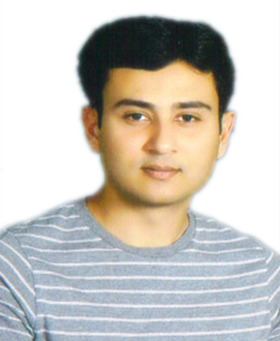}}]{Fahad Shahbaz Khan}
is an Associate Professor (Universitetslektor and Docent) at Computer Vision Laboratory, Linköping University, Sweden and Lead Scientist at Inception Institute of Artificial Intelligence, UAE. He received the M.Sc. degree in Intelligent Systems Design from Chalmers University of Technology, Sweden and a Ph.D. degree in Computer Vision from Autonomous University of Barcelona, Spain. From 2012 to 2014, he was post doctoral fellow at Computer Vision Laboratory, Linkoping University, Sweden. From 2014 to 2018, he was research fellow at Computer Vision Laboratory, Linkoping University, Sweden.  His research interests are in object recognition, action recognition and visual tracking. He has published articles in high-impact computer vision journals and conferences in these areas.

\end{IEEEbiography}


%\vfill


\end{document}